\documentclass{article}

\usepackage{arxiv}

\usepackage[utf8]{inputenc} % allow utf-8 input
\usepackage[T1]{fontenc}    % use 8-bit T1 fonts
\usepackage{hyperref}       % hyperlinks
\usepackage{url}            % simple URL typesetting
\usepackage{booktabs}       % professional-quality tables
\usepackage{amsfonts}       % blackboard math symbols
\usepackage{nicefrac}       % compact symbols for 1/2, etc.
\usepackage{microtype}      % microtypography
\usepackage{lipsum}

\usepackage{appendix}
\usepackage{amsmath}
\usepackage{graphicx}
\usepackage{caption}
\usepackage{subcaption}
\usepackage{siunitx}

\title{A review on deep learning techniques for 3D sensed data classification}

\author{
  David Griffiths\thanks{Corresponding author.} \\
  University College London\\
  London, UK \\
  \texttt{david.griffiths.16@ucl.ac.uk} \\
   \And
 Jan Boehm \\
  University College London\\
  London, UK \\
  \texttt{j.boehm@ucl.ac.uk} \\
}

\begin{document}
\maketitle

\begin{abstract}
Over the past decade deep learning has driven progress in 2D image understanding. Despite these advancements, techniques for automatic 3D sensed data understanding, such as point clouds, is comparatively immature. However, with a range of important applications from indoor robotics navigation to national scale remote sensing there is a high demand for algorithms that can learn to automatically understand and classify 3D sensed data. In this paper we review the current state-of-the-art deep learning architectures for processing unstructured Euclidean data. We begin by addressing the background concepts and traditional methodologies. We review the current main approaches including; RGB-D, multi-view, volumetric and fully end-to-end architecture designs. Datasets for each category are documented and explained. Finally, we give a detailed discussion about the future of deep learning for 3D sensed data, using literature to justify the areas where future research would be most valuable.
\end{abstract}


\section{Introduction}

Extraction of meaningful information from three-dimensional (3D) sensed data is a fundamental challenge in the field of computer vision. Much like with two-dimensional (2D) image understanding, 3D understanding has greatly benefited from the current technological surge in the field of machine learning. With applications ranging from remote sensing, mapping, monitoring, city modelling, autonomous-driving, virtual/augmented reality, and robotics, it is clear why robust and autonomous information extraction from 3D data is in high demand. As such, both academic and industrial organisations are undertaking extensive research projects to further develop this very active field. Classic machine learning methods such as Support Vector Machines (SVM) and Random Forest (RF) have typically relied on a range of hand crafted shape descriptors (i.e. Local Surface Patches \cite{ChenBhanuH2007}, Spin Images \cite{JohnsonHebertA1999}, Intrinsic Shape Signatures \cite{ZhongY2009}, Heat Kernel Signatures \cite{SunEtAlJ2009} etc.) as feature vectors from which to learn. These methods have delivered successful results in a range of 3D object categorisation and recognition tasks \cite{MateiEtAlB2006, ShangGreenspanL2010, GuoEtAlY2013}. However, much like in the field of 2D image understanding there has been a shift in focus to a deep learning approach \cite{LeCunEtAlY2015}. Deep learning approaches differ to other machine learning approaches in that features themselves are learned as part of the training process. This process is commonly referred to as representation learning, where raw data is passed into the learning algorithm and the representations required for detection or classification are automatically derived. This ability to learn features is often seen as the cause for the rapid improvement in 2D and 3D understanding benchmark results. In this paper we offer a comprehensive review of both the foundational and state-of-the-art deep learning methods for classification, object detection and point-wise segmentation of 3D sensed data.

The remainder of this paper is organised as follows; Section 2 looks into the background problem of 3D classification, object detection and segmentation and underlying deep learning architectures developed to allow for state-of-the-art results. Section 3 outlines 3D benchmark datasets. In Sections 4-8 we assess in turn the most common techniques for deep learning of 3D data. Section 9 offers a further discussion of each technique and addresses future research needs. Finally, we conclude the research presented in this paper in Section 10.

\section{Background concepts}

\subsection{Point cloud processing}

With its roots in photogrammetry and more recently lidar, the history of 3D data acquisition is as old as that of 2D image acquisition. In its simplest form, 3D data can be represented in the form of a point cloud. A point cloud is a set of 3D data points $\mathbf{P}\in\mathbb{R}^{3}$ where each point is represented by three coordinates in a Cartesian, or another suitable coordinate system. Additional information can also be included such as colour, and if acquired using active sensors, intensity or reflectance. Unlike 2D images which are stored in structured arrays which provide explicit neighbourhood relations, point clouds in the general case have only implicit neighbourhood relations and are therefore considered unstructured data. Due to the lack of explicit neighbourhood relations point cloud segmentation is a distinct and comprehensive research field.

To assign individual labels to each point in a point cloud a supervised learning approach is usually chosen. Typical supervised learning algorithms include Support Vector Machines (SVM), Random Forest (RF) and Naive Bayes. However, as each point within a point cloud is simply a 3D vector in a Euclidean space, no meaningful feature descriptor can be obtained when only an individual point is considered. To overcome this a local neighbourhood of either $k$ nearest points, or points within a radius $r$ is selected for the respective point and features are computed from the neighbourhood region. As point clouds can contain millions of points, brute force methods to find the local neighbourhood of a given point are computationally expensive and time consuming. Instead accelerated 3D search structures are used such as kd-trees \cite{bentley1975} or approximations thereof (e.g. FLANN \cite{MujaLoweM2009}). Many studies have addressed the issue of neighbourhood size and its dependence of point cloud density \cite{WeinmannEtAlM2015, NiemeyerEtAlJ2014}. Furthermore, to improve this reliance, multiple approaches have extracted neighbourhoods at various scale-spaces to retrieve the optimal neighbourhood size \cite{PaulyEtAlM2003, BroduLagueN2012, DemantkeEtAlJ2012}, thereby providing more robust geometric feature descriptors. To compute the actual features a large range of descriptors have been proposed. Typical recurring features include computing planarity, linearity, scatter, surface variance, vertical range, point colour, eigenentropy and omnivariance \cite{WeinmannEtAlM2015, BeckerEtAlC2017} of the local neighbourhood. Although approaches employing these methods can achieve a high classification accuracy (~90\%), these are generally tested over small areas with little inter class ambiguity. Moreover, these approaches generally do not perform well when point density is sparse or objects are occluded and overlapping.

One common approach to avoid processing in 3D is to rasterise the 3D data from a given perspective centre and viewing angle into a 2.5D structured image array. This allows standard image processing techniques to be employed for segmentation and classification. Moreover, this approach is not restrained to point clouds, given a perspective centre and viewing angle, all forms of 3D data (i.e. meshes, triangulated irregular networks) can be rasterised into 2.5D images. Early work by \cite{HaalaEtAlN1998, HaalaBrennerN1999} made use of such approaches for ALS data extraction of buildings and roads with promising results. This method showed particular promise for extraction of terrain models from digital surface models. \cite{VosselmanG2000a} demonstrated that a slope based approach that closely resembled morphological operators was effective, similarly, \cite{WackWimmerR2002} used a Laplacian of Gaussian operator to detect above ground points after a threshold function is applied to the digital surface model. 

RGB-D has experienced a surge of research credited to the invention of low cost RGB-D sensors such as the Microsoft Kinect{\small \textregistered} and PrimeSense{\small \textregistered}. With their combination of active and passive sensors aiding real time stereo photogrammetry, acquiring (albeit relatively coarse) RGB-D data has become effortless. Whilst these sensors were largely developed by the gaming industry to enable full body console control, they have been used extensively in the field of robotics for navigation, fuelling algorithm development for semantic segmentation. Research regarding low cost sensors can generally be split into the following sub-categories; object tracking and recognition, human activity analysis, hand gesture analysis and indoor 3D mapping \cite{HanEtAlJ2013}. Examples of traditional computer vision processing pipelines include; Chamfer distance matching \cite{XiaEtAlL2011}, depth-based graph-cut object segmentation \cite{yin2013hierarchical} and super-voxel based \cite{AijaziA2013}. The reader is referred to \cite{AijaziA2013} for a full review on RGB-D object recognition.

\subsection{Deep learning processing}

Although this review paper is concerned with the segmentation of 3D data, it is important to acknowledge the progression of Convolutional Neural Networks (CNN) for 2D image understanding. The progress made with 2D images acts as a foundation for which many 3D learning algorithms build upon. CNNs in general tend to deal with the task of classification of a structured array. Common applications are the classification of a whole image (classification), a localised subset of the image (object detection), individual pixels (segmentation), or individual pixels that relate to single objects (instance segmentation). Here we will look at some of the most influential and ground breaking CNN architectures which have helped shape the success of CNNs. It should be made clear that object detection and segmentation are effectively a fine classification inferences and therefore, classification papers are equally relevant for object detection and segmentation tasks alike.

Often referred to as 'the one that started it all', AlexNet \cite{KrizhevskyA2012} was the pioneering network architecture that won the 2012 ImageNet Large-Scale Visual Recognition Challenge (ILSVRC) with a TOP-5 test accuracy of 84.6\%. Whilst this unofficial title does perhaps divert due credit to the original CNN paper \cite{LeCunEtAlY1999}, there is no denying the wave of attention AlexNet brought to CNNs for image classification. Whilst AlexNet was a deep network it was comparatively shallow to more current network architectures. A subsequent key landmark development in classification is the concept of residual blocks proposed in Microsoft's ResNet \cite{HeEtAlK2016} architecture. ResNet was not only notably deep (152 layers) but introduced a distinct method to allow for training of such deep networks. It achieved this with identity skip connections. These connections allow the layers to copy their inputs into later layers. Say we have an input $x$, and $F(x)$ is the result of $x$ after it has been passed through a convolutional-ReLU-convolutional sequence. The identity of $x$ is then added to $F(x)$ given the output as $F(x)+x$. This means the next layer learns new information but helps retain initial knowledge learned in prior layers. These additional operations allow the gradient to flow more freely through the graph during back propagation which helps overcome the vanishing gradients problem.

There has been much deliberation and experimentation on how best to use coarse whole image classification networks for localisation of specific objects present within the scene. The simplest approach to this problem is to apply a dense sample of classifications over an input image and record the bounding box coordinates of the areas with the highest prediction probabilities. Although the quality of the results obtained in this manner can be high, inference is computationally expensive and real time processing unrealistic. Generally, the proposed solutions to this problem can be split into two categories; one-stage and two-stage object detectors. The two-stage approach was popularised by \cite{GirshickEtAlR2014} and, put simply, works by first generating a region proposal for potential bounding box locations, and secondly, classifies each region proposal candidate using a classification CNN (i.e. ResNet \cite{HeEtAlK2016}). The one-stage approach was popularised by  \cite{sermanet2013overfeat, liu2016ssd, redmon2016you} and motivated by the potential to speed up the two-stage process which has many speed limitations such as their inability to optimise or parallelise. Instead, the one-stage detector applies a dense sample of classifications over the image at various scales and aspect ratios. The classifications with the highest probability scores for a given object are used as the object's location within the image. This computationally cheaper method has allowed for one-stage object detectors to be deployed on fairly basic hardware (i.e. mobile phones) for real time detection, however, usually achieve poorer accuracy than their two-stage counter-parts. Despite this, it is evident that the gap between one-stage and two-stage detectors is narrowing.

The cornerstone of most modern CNN segmentation networks is the Fully Convolutional Network (FCN) \cite{LongJ2015}. FCN takes an existing classifier and removes the final fully connected layers (known as the encoder). Next, convolutional transpose\footnote{Also referred to as deconvolutional, backward strided convolution, upconvolution and fractionally strided convolutions in literature.} layers are included in the network architecture. Convolutional transpose operators have the inverse effect of a convolution layer and upsample the intermediate tensors until they match the original size of the image (known as the decoder). As convolutional transpose layers are a form of convolution layers, their weights can be learned during backpropagation. One key issue with the initial network is that during the convolution layers much of the global spatial information is lost. To account for this, skip connections are introduced to combine spatial information from earlier layers, producing a finer segmentation. Despite being a seminal paper, a few short comings exist for FCNs. Firstly, even with skip connections a large amount of global spatial information is lost. Secondly, individual instance awareness is not possible. Lastly, the network is not very efficient when high resolution images are inputted, and therefore, real time applications are not viable. Since the original FCN paper, a range of architectures have been published offering improvements. One of the earliest prominent papers was SegNet \cite{BadrinarayananEtAlV2017}. SegNet created a decoder where each upsampling layer corresponds to the equivalent max pooling layer in the encoder. The feature maps are than upsampled with the max pooling indices. Once the full resolution is restored, a softmax classifier is used to perform pixel-wise segmentation. A plethora of FCN derived CNNs have subsequently been proposed.

To enable instance segmentation, \cite{HariharanEtAlB2014} proposed a method where first an object detector network is performed on the image to find each object instance. Their pipeline also uses the multi-scale combinatorial grouping \cite{ArbelaezEtAlP2014} to compute region proposals for potential objects. This was built on top of a Region-CNN \cite{GirshickEtAlR2014} and classified each region using a linear SVM. \cite{PinheiroEtAlP2015} developed a fully end-to-end object detection and segmentation network DeepMask, which allows for real time instance segmentation inference. Here the model predicts a segmentation mask and the likelihood of the mask containing an object. This was further refined resulting in the network SharpMask \cite{PinheiroEtAlP2016a}. SharpMask refines the coarse masks produced by DeepMask by utilising features from progressively earlier layers in the network. Mask R-CNN \cite{HeEtAlK2017a} builds directly on Faster R-CNN by adding a third output branch for each candidate object. Whereas the prior branches were responsible for class label and bounding box offset, the authors now output object mask proposals. This can therefore be trained in parallel with the Faster R-CNN network, outputting a binary mask for each Region of Interest (RoI). The loss function is updated to then include the mask predictions so that; $\mathcal{L}=\mathcal{L}_{cls}+\mathcal{L}_{box}+\mathcal{L}_{mask}$, where $\mathcal{L}$ is the total loss.

\section{Benchmark datasets} \label{datasets}

Benchmark datasets have a long tradition in computer vision, robotics and geospatial communities. Not only do benchmarks allow for fair comparison between various algorithms/approaches, but they also provide the community with free to use high quality training data. This is particularly prominent in the field of deep learning where large datasets are required for network training. Generally, benchmark datasets are created by large industry or university research groups. Naturally, the availability, diversity and size of benchmarks decrease as complexity increases. For example, single image classification (i.e. dog, cat, car etc.) requires very low user skill to be quickly classified. Object detection requires more time and skill and is more subjective. Segmentation requires patience and is again more subjective. Complexity and time of dataset labelling arguably increases exponentially with dimensions and purpose (i.e. classification, object detection, segmentation). 3D segmentation is therefore a complex procedure which requires a skilled user, patience and an acute eye for detail. For this reason, large scale, accurately labelled 3D segmentation benchmark datasets are sparse. Labelling RGB-D data is an easier task as this can be performed in a 2D space and the classes propagated into the depth dimension. In this section we describe popular RGB-D and 3D publicly available datasets. We present datasets in chronological order as per category.

\subsection{RGB-D datasets}

\begin{itemize}
	
	\item \textbf{RGB-D Object Dataset} \cite{LaiEtAlK2011a}\footnote{https://rgbd-dataset.cs.washington.edu}: The RGB-D object dataset was developed by researchers at Washington University, USA. The dataset consists of 11,427 manually segmented RGB-D images. The images contain 300 common objects which have be classified into 51 classes arranged using WordNet hypernym-hyponym relationships (similar to ImageNet). The images are acquired using a \textit{"Kinect style"} sensor to generate 640 x 480 RGB-D frames and a frequency of 30\textit{H}$z$. 22 validation video sequences are also provided for evaluating performances.

  \item \textbf{NYUDv2} \cite{SilbermanEtAlN2012}\footnote{https://cs.nyu.edu/~silberman/datasets/nyu\_depth\_v2.html}: The New York University Depth Dataset v2 (NYUDv2) was developed by the researchers at the New York University, USA. The dataset consists of 1449 RGB-D segmentation labels for images of indoor scenes. The dataset was captured using the Microsoft Kinect v1. The objects are split into 40 classes and 407,024 validation images are provided. The dataset is mainly aimed at aiding training for robotics navigation applications.

  \item \textbf{SUN3D} \cite{XiaoEtAlJ2013a}\footnote{http://sun3d.cs.princeton.edu}: Developed at Princeton University, USA, the SUN3D dataset contains 415 RGB-D sequences captured using an Asus Xtion sensor. The scenes vary across 254 indoor spaces. Each frame has been manually segmented to object level. As the dataset is reconstructed using Structure-from-Motion (SfM) camera pose for each scene is also available. The researchers open sourced their labelling tool to help aid research.
  
  \item \textbf{ViDRILO} \cite{Martinez-GomezEtAlJ2015}\footnote{http://www.rovit.ua.es/dataset/vidrilo}: The Visual and Depth Robot Indoor Localization with Objects information (ViDRILO) dataset was developed by researchers at the Universidad de Castilla-La Mancha and Universidad de Alicante, Spain and funded by a Spanish government incentive. The dataset contains 22,454 RGB-D segmentation images captured over five indoor scenes using a Microsoft Kinect v1 sensor. The data is acquired under challenging lighting conditions. Each RGB-D image is labelled with the semantic category of the scene (corridor, professor office, etc.) but also with the presence/absence of a list of pre-defined objects (bench, computer, extinguisher, etc.). The dataset was released to benchmark multiple problems such as; multimodal place classification, object recognition, 3D reconstruction and point cloud data compression.
  
  \item \textbf{SUN RGB-D} \cite{SongEtAlS2015}\footnote{http://rgbd.cs.princeton.edu}: The SUN RGB-D dataset was developed by the same research team as the SUN3D dataset at Princeton. The dataset was motivated by the gap between 3D reconstruction and scene understanding of RGB-D datasets. The dataset was acquired using four sensors; Intel RealSense, Asus Xtion, Microsoft Kinect v1 and Microsoft Kinect v2. The dataset contains 10,000 manually segmented images split into 63 classes of indoor scenes. The dataset also boasts 146,617 2D polygons and 58,657 3D object bounding boxes.
 
\end{itemize}

\subsection{Indoor 3D datasets}

\begin{itemize}
	
	\item \textbf{A Benchmark for 3D Mesh Segmentation} \cite{ChenEtAlX2009}\footnote{http://segeval.cs.princeton.edu}: The benchmark for 3D mesh segmentation was developed by researchers at Princeton University, USA. The dataset contains 380 meshes across 19 common object categories (i.e. table, chair, plane etc.). The dataset segments each mesh into functional parts and is designed to aid research into 3D part semantic segmentation. The dataset is also aimed at aided research into how humans decompose objects into individual meaningful parts. The ground truth is derived from darker lines caused by geometric edges in the mesh.
	
	\item \textbf{ShapeNet} \cite{ChangEtAlA2015} \footnote{https://cs.stanford.edu/~ericyi/project\_page/part\_annotation/}: ShapeNet is a large scale repository for 3D CAD models developed by researchers from Stanford University, Princeton University and the Toyota Technological Institute at Chicago, USA. The repository contains over 300M models with 220,000 classified into 3,135 classes arranged using WordNet hypernym-hyponym relationships. ShapeNet consists of multiple subsets however, the most relevant for segmentation is the ShapeNet Parts subset \cite{YiEtAlL2016}. ShapeNet Parts contains 31,693 meshes categorised into 16 common object classes (i.e. table, chair, plane etc.). Each shapes ground truth contains 2-5 parts (with a total of 50 part classes). Ground truth points are derived from regular point sampling from the mesh.
	
	\item \textbf{Stanford 2D-3D-Semantics} \cite{ArmeniEtAlI2017}\footnote{http://buildingparser.stanford.edu/dataset.html}: The Stanford 2-D-S dataset was developed by researchers in Stanford University, USA. The dataset is made up of a variety of mutually registered modalities from 2D (RGB), 2.5D(RGB-D) and 3D domains. The dataset contains over 70,496 images, all of which have corresponding depth, surface normal directions, global $x,y,z$ positions and semantic segmentation labels (both per-pixel and per-point). Each pixel and point is categorised into 13 classes; ceiling, floor, wall, column, beam, window, door, table, chair, bookcase, sofa, board and clutter. The scenes vary over six large scale indoor areas from three educational/office buildings covering 271 rooms. All data is referenced into the same coordinate reference system, which reduces 70,496 1080x1080 RGB images to 1,413 equirectangular RGB images. A Matterport 3D camera is used to compute approximately 700M depth measurements.
	
	\item \textbf{ScanNet} \cite{DaiEtAlA2017a}\footnote{http://www.scan-net.org/} : ScanNet is a RGB-D video dataset which contains 2.5M views in $>$1500 scans. All scans are reconstructed into 3D mesh models using BundleFusion \cite{dai2017bundlefusion}. Semantic segmentation was performed by crowd sourcing using a novel labelling application developed by the authors. Although the dataset is not a point cloud dataset by default, point clouds can be extracted by sampling the mesh for each scene or extracting mesh vertices. Typically, each scene contains ~400-600k points. This can be doing either uniformly or non-uniformly, and metrics are often presented for both. The dataset contains 20 object classes commonly found in residential accommodations such as wall, floor, window, desk, picture etc.

\end{itemize}

\subsection{Outdoor 3D datasets}

\begin{itemize}
	
	\item \textbf{Oakland} \cite{MunozEtAlD2009}\footnote{http://www.cs.cmu.edu/~vmr/datasets/oakland\_3d/cvpr09/doc}: The Oakland 3D point cloud dataset was developed by researchers at the Carnegie Mellon University, USA. The dataset covers an outdoor section of the Carnegie Mellon University campus in Oakland, Pittsburgh. The data was acquired using car mounted a MLS with a SICK LMS lidar sensor using a push broom sensor for range measurements. The complete dataset has 1.6M manually classified points. The dataset was remapped from originally 44 to 11 classes. The classes include common outdoor, road side objects such as; ground, facade wall and column, post, tree trunk, shrub etc.
	
	\item \textbf{Sydney Urban Objects} \cite{QuadrosEtAlA2012}\footnote{http://www.acfr.usyd.edu.au/papers/SydneyUrbanObjectsDataset.shtml}: The Sydney Urban Objects dataset was developed by researchers at the University of Sydney, Australia. The data was captured using a car mounted MLS with a Velodyne HDL-64E lidar sensor for range measurements. The data is acquired around the Sydney area and contains 2.3M manually classified points. The points range 26 classes of common road and street objects. The $x,y,z$ data is also accompanied with time, azimuth, range and intensity data. Although referred to as a MLS dataset, no trajectory information is given with the scans .csv files and therefore each scan is effectively an unregistered static scan. The dataset does however also come with .bin files which require an open source library (developed by the same group) for viewing. Due to compilation issues we were not able to test if these files contain trajectory information. The dataset also provides per object scans (i.e. pre-segmented to object level). This allows for testing of whole point cloud classification as well as per-pixel semantic segmentation. 
	
	\item \textbf{Paris-rue-Madame} \cite{SernaEtAlA2014}\footnote{http://cmm.ensmp.fr/~serna/rueMadameDataset.html}: The Paris-rue-Madame dataset was developed as part of the TerraMobilita project and Mines ParisTech. The dataset consists of a 160M MLS scan of rue Madame (street), 6th Parisian district, France. The range measurements were recorded with a Velodyne HDL-32 lidar scanner mounted onto the StereopolisII mobile mapping system. A total of 20M points are collected, storing, $x,y,z$ and reflectance. All $x,y,z$ points are georeferenced into a geographical coordinate system. 17 classes are categorised which are very similar to other road side MLS datasets described. 2.3M points are recorded however many are stray points from windows in facades, causing a significant proportion of points to be unusable and detrimental for model training.
	
	\item \textbf{iQmulus/TerraMobilita} \cite{ValletEtAlB2015}\footnote{http://data.ign.fr/benchmarks/UrbanAnalysis}: Similar to the Paris-rue-Madame, the iQmulus / TerraMobilita dataset was funded by various projects including; iQmulus, TerraMobilita along with the Institut National De L'Information G\'eographique et For\'estiere (IGN) and Mines ParisTech, France, along with several European universities. The data is acquired with the StereopolisII MLS using a car mounted \ang{360} Riegl LMS-Q120i sensor for range measurements. The use of the Riegl scanner results in a much higher point density when compared to other datasets using Velodyne scanners. The dataset contains approximately 300M points. Each point also records time, reflectance and number of echoes. All $x,y,z$ points are georeferenced into a geographical coordinate system. The class tree is detailed with 50 classes of common road and street objects sorted in a hierarchical manner. Although the dataset does have a diverse range of classes, a large variance in the number of points per class exists. The dataset is also segmented in 2D space in a semi-manual approach. This does result in poor classification accuracy in complex areas.
	
	\item \textbf{TUM City Campus} \cite{GehrungEtAlJ2017}\footnote{https://www.iosb.fraunhofer.de/servlet/is/71820}: TUM City Campus dataset was developed by Fraunhofer IOSB at the Technical University of Munich, Germany. Similar to the Sydney Urban Objects dataset, the TUM City Campus dataset consists of a car mounted MLS with a Velodyne HDL-64E sensor for range measurements. All $x,y,z$ points are georeferenced into a local Euclidean coordinate reference system. Sensor position and reflectance are also recorded. All points are stored in \textit{PCD} format which is the native format for the Point Cloud Library, a popular point cloud processing library written in c++. Nine classes are defined; artificial terrain, natural terrain, high vegetation, low vegetation, building, hardscape, scanner artefact and unclassified. Over 8000 scans were acquired with the sensor resulting in over 1.7BN points. However, only a small subset of these points have been manually classified. 
	
	\item \textbf{Semantic3D.NET} \cite{HackelEtAlT2017}\footnote{http://semantic3d.net}: Semantic3D.NET was developed by researchers at ETH Zurich, Switzerland. The dataset was created with two fundamental motivations. Firstly, all of the prior benchmarks and datasets discussed are forms of MLS datasets. Semantic3D.net is instead a large network of static TLS scans. Secondly, the above outdoor datasets are arguably not large enough to train deep learning network architectures. To account for this the total Semantic3D.net dataset contains approximately 4BN points collected over 30 non-overlapping scans. The points are classified into eight classes; man made terrain, natural terrain, high vegetation, low vegetation, buildings, hardscape, scanning artefacts and cars. The scans also cover various scene types including; urban, sub-urban and rural. The competition page contains the current leaders along with information for entry. The standardised evaluation metrics are defined in the release paper. Although not explicit, there is a strong emphasis on deep learning within this dataset.
	
	\item \textbf{Paris-Lille-3D} \cite{roynard2018paris}\footnote{http://npm3d.fr/paris-lille-3d}: Created by researchers at Mines ParisTech, Paris-Lille-3D is an urban MLS dataset containing 143.1M labelled points covering 50 classes. The data was generated using a Velodyne HDL-32E scanner positioned near-perpendicular to the road surface (in comparison, for autonomous vehicles Velodyne scanners are mounted parallel to the road surface). The entire dataset is comprised of three subsets consisting of 71.3M, 26.8M and 45.7M points.
	
\end{itemize}

\section{RGB-D}\label{RGB-D}

Although not restricted by definition, the majority of RGB-D methods proposed in literature are designed around indoor scenes. The reason for this is these studies generally use a low cost active RGB-D sensor for data acquisition. Such sensors rely on infrared light to project a pseudo-random dot matrix for stereo photogrammetry. The systems do not therefore fare well in outdoor environments where strong infrared noise is present. Furthermore, as RGB-D sensors generally have a small baseline they are only suitable for short range navigation. This makes them unsuitable for a large range of outdoor navigation problems. In this section we look at deep learning approaches to object detection and semantic/instance segmentation of RGB-D data.

\begin{figure}[ht!]
\begin{center}
	\includegraphics[width=0.75\textwidth]{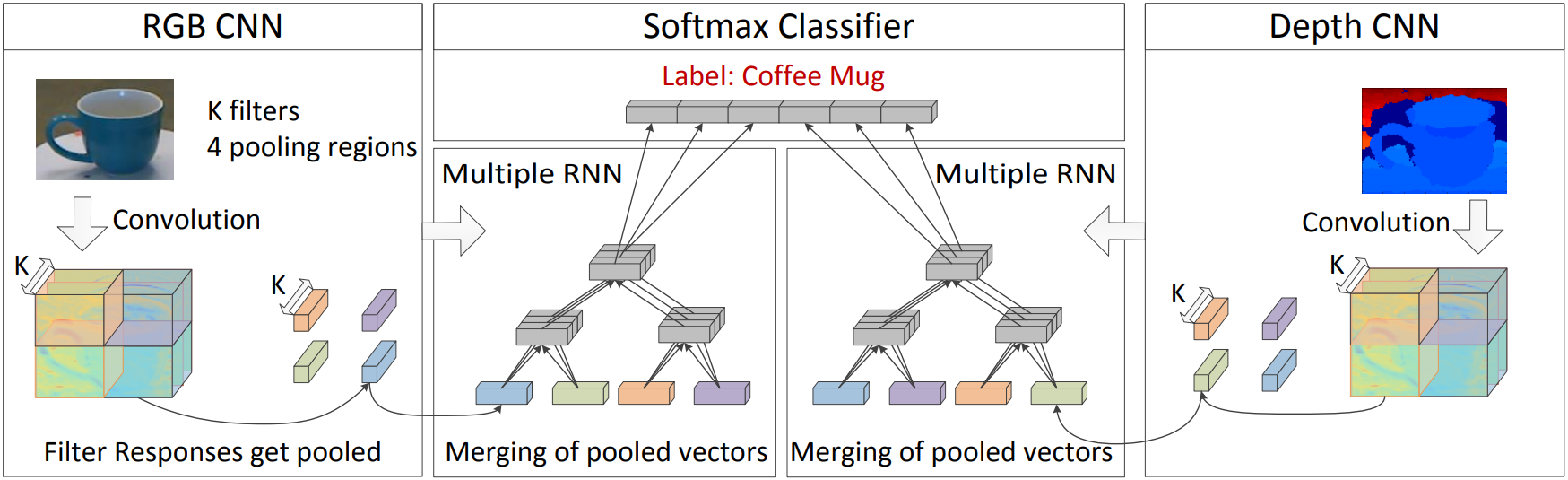}
	\caption{Network architecture to process RGB and depth images separately to learn low level features. Multiple RNNs are then used to find higher level features. Finally feature vectors are concatenated for a final softmax classification layer. Image source: \cite{socher2012convolutional}}
	\label{fig:cnn_rnn_socher}
\end{center}
\end{figure}

One of the earlier methods for classifying RGB-D is a combination of a CNN and Recurrent Neural Network (RNN) \cite{socher2012convolutional}. The approach works by first learning low level translation invariant features with CNNs for both an RGB and corresponding depth image independently. The learnt features are then passed into multiple, fixed tree RNNs to generate more high level global features. RNNs were used here as they were seen to reduce convolutional and pooling functions into a single efficient hierarchical operation, resulting in improvements with regard to computation time (Fig. \ref{fig:cnn_rnn_socher}). A similar approach was also taken by \cite{EitelEtAlA2015} who used two CNNs operating on RGB and depth independently. The two networks are then fused together in a fully connected layer after two fully connected layers for each stream respectively. The network can therefore learn weights with respect to the fusion of the two independent streams as well as extracting individual features from each input. \cite{CouprieEtAlC2013} proposed at the time the first full scene segmentation using deep learning by adapting previous work by \cite{FarabetEtAlC2013}. In this method the RGB and depth channel are passed into a Laplacian pyramid to extract features at three scales. The features are fed into separate CNNs to generate feature maps which are then concatenated into a single feature map. In parallel a single segmentation of the image into superpixels is performed, exploiting the natural contours of the image. The CNN derived feature maps and superpixels are then aggregated into a single prediction of each pixel. This method demonstrated the potential of a 4-channel CNN architecture, removing the need to independent RGB and depth image processing as seen in \cite{socher2012convolutional}.

\begin{figure}[ht!]
\begin{center}
	\includegraphics[width=0.75\textwidth]{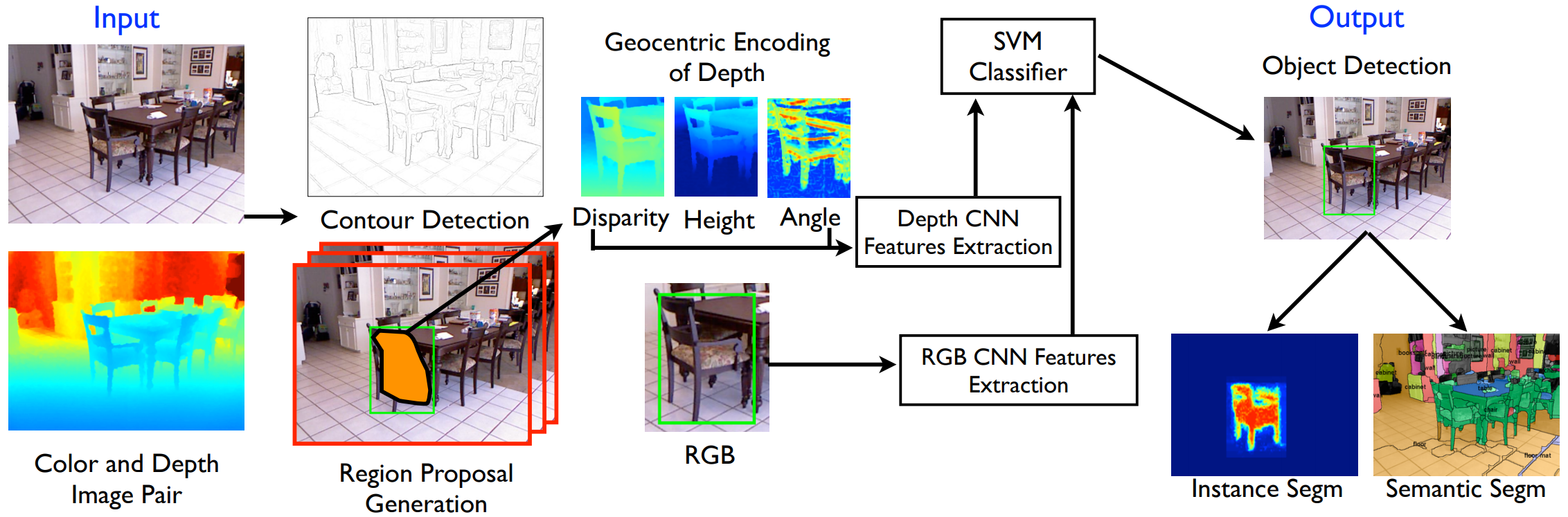}
	\caption{Pipeline for object detection and instance segmentation of RGB-D images. First a random forest classifier is used for contour extraction and a random forest regressor used for region proposals. HHA image channels are computed for the depth image and features extracted using a CNN. Simultaneously, RGB features are extracted using a CNN. All features are passed into a SVM for final classification. Foreground background segmentation is then applied for instance segmentation. Image source: \cite{gupta2014learning}}
	\label{fig:rich_features}
\end{center}
\end{figure}

Although simply stacking a depth channel onto an existing CNN architecture is a valid method, this does not fully exploit the geometric information encoded in the depth channel. \cite{gupta2014learning} explored this by representing each pixel in terms of horizontal disparity, pixel height above ground and angle between normals and gravity (known as HHA). These three computations are then stacked into a three channel image. Extraction of features using CNNs from HHA channels were demonstrated to learn stronger representations and achieved high performances. The pipeline first implements a structured forest approach \cite{DollarZitnickP2013, GuptaEtAlS2013} to extract contours from the RGB-D image. This is then used to generate region proposals in the depth channel of the image using a RF regressor. The HHA of the region proposal is generated and passed into a CNN for feature extraction. Simultaneously, the RGB channels of the region proposal are processed through a CNN for feature extraction. Both the RGB and depth channel features are subsequently passed into a SVM classifier. Interestingly, this approach can then perform instance segmentation by performing a foreground background segmentation algorithm (Fig. \ref{fig:rich_features}). This was tested on a modified R-CNN \cite{GirshickEtAlR2014} object detection architecture and demonstrated a 54\% relative improvement over the (at the time) state-of-the-art methods. Adapting a similar methodology to ALS data for land classification, \cite{balado2018automatic} generate raster maps where red, green and blue values for individual pixels are represented by the height difference, medium intensity and number of returns of all points within the pixel area respectively. This offers a more feature rich raster map which is then passed into a CNN for feature extraction and segmentation. Although this method does not demonstrate state-of-the-art results, it demonstrates a sound methodology for alternative raster map representations for ALS point clouds.

\cite{LiEtAlZ2016} proposed a Long-Short-Term-Memory (LTSM) Fusion on top of multiple convolutional layers to fuse RGB and depth data. Depth data is first converted into HHA channels before being passed into convolutional layers for feature extraction. RGB data is simultaneously passed through a DeepLab CNN \cite{ChenEtAlL2014}. The two sets of features are then fed into respective memorised context layers to model both short-range and long-range spatial dependencies along the vertical direction. The contexts are then integrated using a memorised fusion layer from both RGB and depth channels. A bi-directional propagation of the fused vertical contexts give a 2D global context of the image. This approach at the time achieved state-of-the-art results on the SUN RGB-D and NYUDv2 dataset with a mean accuracy of 48.1\% and 49.4\% respectively. \cite{HazirbasEtAlC2017} demonstrated how more simple encwith the network architecture FuseNet. The network architecture is similar to that of SegNet however, instead of a single decoder network, two decoder networks are constructed for RGB and depth respectively. Each layer of the two decoder networks is fused after each pooling function (i.e. $\mathcal{F}_{rgb}\oplus \mathcal{F}_{depth}$). As a preprocessing step the depth channel is normalised to the RGB image ([0, 255]) to ensure the same colour range. The fusion block is an element wise summation of the two decoder layer weights. The inclusion of the depth data feature maps is shown to enhance the RGB feature maps, resulting in geometrically aware RGB feature maps. FuseNet obtained a 76.27\% accuracy on the SUN RGB-D benchmark, substantially higher than preceding submissions. 

Alternative approaches make use of multi-vew RGB-D data. Whilst such data is only possible in relatively small contained areas, it has attracted a lot of attention in the field of warehouse robotics where autonomous \textit{pick-and-place} systems are sought. \cite{ZengEtAlA2016} propose a method where a robotic arm is used to acquires multi-view RGB-D images which are subsequently fed into a FCN. Both AlexNet and VGG-16 were used as classifiers for the FCN, however, concluded the depth information did not yield any significant improvements in the overall segmentation result. While there are numerous variables that could have caused this (i.e. high noise levels in the depth image), this could suggest, in line with \cite{gupta2014learning}, that simply stacking a depth channel onto 3-channel CNN is not effective. \cite{MaEtAlL2017} further justify the preceding statement by applying a multi-view RGB-D approach to a FuseNet encoder-decoder type network. To enforce consistency for training a RGB-D SLAM is applied to frames of the moving camera to determine a 6-D camera pose and trajectory. One of the main contributions of the paper is the enhancement of FuseNet with a multi-scale loss minimisation. The network also differs by fusing the first three convolutional layers then having two RGB-D convolutional layers prior to the decoder layers. Furthermore, semantic segmentation predictions from multiple frames utilise trajectory information to fuse into one single key frame. This approach achieves state-of-the-art in both single view semantic segmentation NYUDc2 benchmarks where a pixel-wise accuracy of 79.13\% and 70.66\% for the 13 and 40 class segmentation challenges respectively. 
 
\section{Volumetric approaches}

A common approach for extracting features directly on 3D data is to represent shapes as volumetrically discretized (i.e. voxel) data. By representing data in this form, 3D convolutions can be used to extract meaningful information. A 3D convolution operator functions in a similar manner to a 2D convolution operator however instead takes a shape $w,h,l$ as input, applies a kernel $k,k,d$ and outputs shape $w,h,m$. Each stride traverses the grid similar to a 2D convolution operator and increases depth by a given stride once each 2D plane has been convolved. Formally, given a 3D convolution operator $C(n,d,f)$ where input is $n\times n\times n$ and $d$ feature maps of size $f\times f\times f$. The output at position $x,y,z$ on the $m$-th feature map of layer $l$ is:

\begin{equation}
\nu^{x,y,z}_{lm} = b_{lm}+\sum_{q}\sum_{i=0}^{f-1}\sum_{j=0}^{f-1}\sum_{k=0}^{f-1}w^{ijk}_{lmq}\nu^{(x+i)(y+j)(z+k)}_{(l-1)q}
\end{equation}

\noindent where $b_{lm}$ is the layer bias, $q$ goes through the feature maps in the $l-1$-th layer, $w^{ijk}_{lmq}$ are the weights at position $i,j,k$ of the kernel of the $q$-th feature map. 

One of the first significant successes of volumetric CNNs for 3D object classification was proposed by \cite{MaturanaSchererD2015} with the network architecture VoxNet. VoxNet is a shallow and comparatively basic 3D-CNN model architecture that demonstrated the potential for 3D convolution operators to learn features from voxel occupancy grids. Occupancy grids were used as they maintain a probabilistic estimation of free, occupied and unknown space from range measurements. This varies from point cloud data where only unknown and occupied space is represented, hence free and unknown space is distinguished. A prerequisite for VoxNet is a pre segmented point cloud subset for which a single object classification is derived. This makes VoxNet unsuitable for segmentation, although object detection is possible with either a sliding window approach or initial segmentation algorithm (i.e. Euclidean clustering etc.). Fixed spatial resolutions of 0.1m$^2$ are used for lidar data and mapped to an occupancy grid of size $32\times32\times32$. For preprocessing, input data is normalised between $[-1, 1]$. The network contains two convolutional layers, a pooling layer, and two fully connected layers. The final model architecture was derived from a stochastic search of hundreds of 3D-CNN architectures on synthetic lidar data for simple classification tasks.

Based on a similar concept to VoxNet, ShapeNets \cite{WuEtAlZ2015} takes as an input a depth map of an object acquired using a low cost sensor (i.e. Kinect) from a single view. The object is then represented as a probabilistic distribution of binary variables on a 3D voxel grid. The volumetric shape is passed into a convolutional deep belief network with 3D convolution operators for feature extraction. The purpose of the model is not only to simply recognise the object but also perform a missing parts reconstruction using a database of known shapes. To help drive research in partial object reconstruction the authors also released the ModelNet public dataset containing 3D Computer Aided Design (CAD) models. As computing features from voxel grids are computationally expensive, a grid size of $24\times24\times24$ with three cells in each dimension for padding was opted for. Although the primary intended application of the network is for partial object reconstruction completion, the network achieved at the time of publication state-of-the-art results on the NYU depth dataset.

A recent advancement on \cite{MaturanaSchererD2015, WuEtAlZ2015} is proposed by \cite{HackelEtAlT2017} to adapt the method for per-point semantic segmentation. Each point is represented as a set of local voxel grids of size $16\times16\times16$ at five resolutions ranging from 2.5cm to 40cm, deriving an input tensor of shape $5\times16\times16\times16$. Empty and filled voxels are encoded with $[0,1]$ respectively. Features are extracted from each resolution independently, concatenated together and passed through two fully connected layers. The feature extraction is similar to \cite{MaturanaSchererD2015}, consisting of three operations of convolutions, ReLU and max pooling. This method achieved an overall accuracy of 77.2\% on the Semantic3D benchmark dataset which was published by the same authors.

\begin{figure}[ht!]
    \centering
	\includegraphics[width=0.75\textwidth]{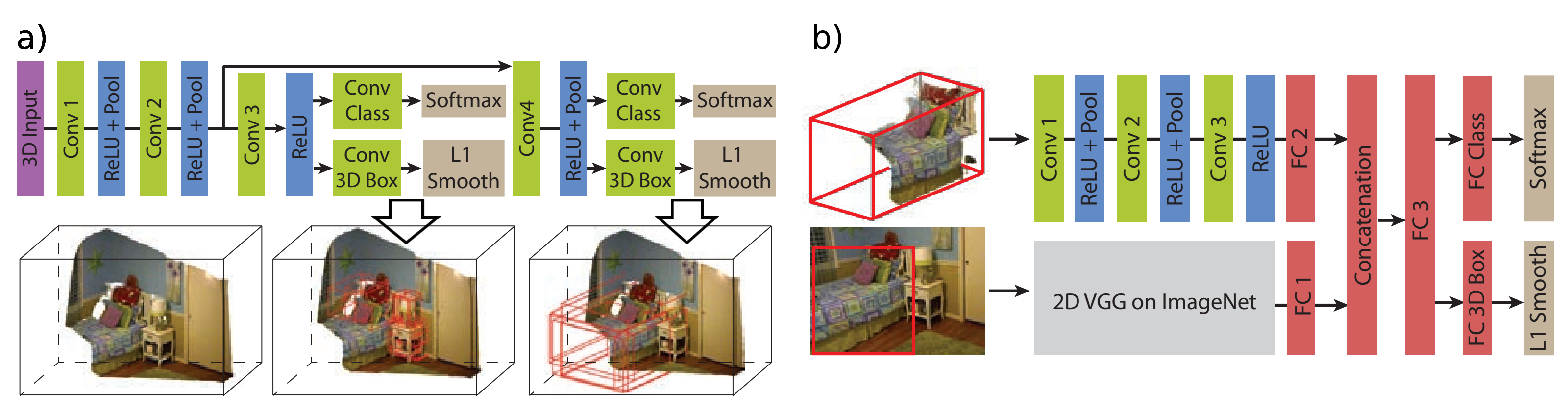}
	\caption{a) 3D amodal region proposal network used for the deep sliding network architecture. Two receptive fields 0.4m$^2$ and 1m$^2$ for small and large objects respectively. b) Joint object recognition network. Features are extracted from both 3D voxel grids and 2D spectral images and concatenated together to exploit features from each feature space. Image source: \cite{SongXiaoS2016}.}
	\label{fig:deep_sliding}
\end{figure}

\cite{SongXiaoS2016} proposed a method to take depth images from low cost sensors and perform amodal 3D object classification and localisation (object detection). Based on sliding shapes \cite{SongXiaoS2014}, which requires hand crafted features trained using a SVM, a fully end-to-end deep learning approach utilising a 3D CNN was proposed. The network takes a similar approach to the 2D Faster R-CNN where both a region proposal network and object detection network are trained simultaneously. For a given indoor 3D scene the major directions of the space are computed using the Manhattan world assumption and is rotated to align with gravity. A uniformly spaced 3D voxel grid then encodes the 3D information with a directional truncated signed distance function. The voxel grid is fed into a multi-scale 3D region proposal network (Fig. \ref{fig:deep_sliding}). To enable multi-scale, two receptive field sizes are defined, 0.4m$^2$ and 1m$^2$ for small and large objects respectively. To make full use of the RGB-D 3D region proposals are projected back to the image plane and a 2D-VGGNet is trained to classify each object, allowing for depth and spectral information to be exploited. The features from each network are concatenated and fed into a fully connected layer, followed by independent fully connected layers for classification and 3D bounding boxes. Results presented by the authors suggested 3D-CNNs are more powerful feature extractors than 2D representations of 3D data.

Despite the above methods performing well on a range of benchmark datasets, volumetric CNNs are generally considered to perform worse than multi-view 2D CNNs (\textit{Section \ref{multi-view}}). The two main variables causing the disparity in results are network architecture design and input resolution. Intuitively, it would be assumed that volumetric data would encode more information than multiple 2D images. \cite{QiEtAlC2016} demonstrated that this is not the case. By using sphere renderings of a 3D voxel grid a state-of-the-art multi-view CNN is trained. Whilst this improved existing volumetric CNN results it still did not perform as well as 2D multi-view renderings of the object. To address this issue \cite{QiEtAlC2016} proposed two new network architectures. The first is designed to mitigate overfitting by introducing auxiliary training tasks. The second architecture is designed to mimic multi-view CNNs by projecting 3D shapes to 2D by convolving its 3D volume with an anisotropic probing kernel.

As overfitting is a key problem when training volumetric CNNs, \cite{QiEtAlC2016} introduce auxiliary training tasks that are similar to the main global task but are difficult to overfit. This ensures learning continues even if the global task has overfit. The tasks involve predicting the class on local sub-volumes of the input. This allows completion of difficult tasks without the requirement of additional annotations. To implement this 3D equivalent \textit{multilayer perceptron convolution} (\textit{mlpconv}) layers \cite{LinEtAlM2013} are used. The network consists of three \textit{mlpconv} layers. The network then branches into two separate classifiers. Whereas one branch takes the whole image for classification following three fully connected layers and a softmax layer, the other branch is designed for auxiliary tasks. The $512\times2\times2$ 4D tensors are sliced into $2\times2\times2=8$ vectors of dimension 512. Each vector then has its own classification task. 

\begin{figure}[ht!]
    \centering
	\includegraphics[width=0.75\textwidth]{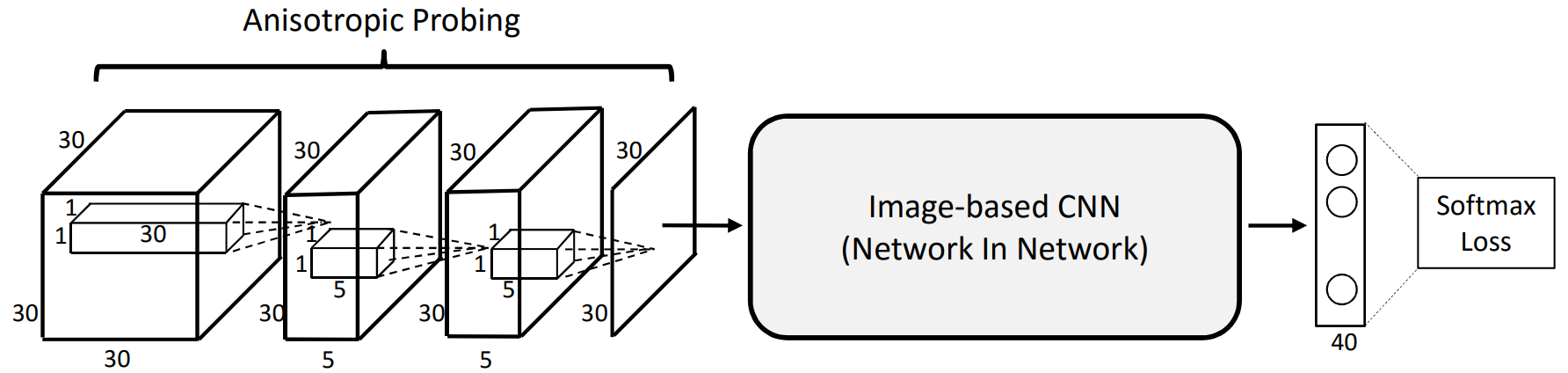}
	\caption{CNN with anisotropic probing kernels. Elongated 3D convolutions are used to first extract features from 3D. The 3D output is subsequently aggregated onto a 2D plane and passed into a 2D Network-in-Network CNN and classified using a softmax layer. Image source: \cite{QiEtAlC2016}.}
	\label{fig:anisotropic}
\end{figure}

As seen with RGB-D (\textit{Section \ref{RGB-D}}) and multi-view networks (\textit{Section \ref{multi-view}}), a common approach for handling 3D data is to project the data into 2D. However, this is often at the cost of some degree of information loss. The second approach by \cite{QiEtAlC2016} tackles this by using anisotropic probing kernels to first extract features in 3D and then project the features into 2D. An elongated kernel is used for three 3D convolutional operations, each followed by a ReLU. Finally, data is aggregated to a 2D plane where it is passed into a image based Network-in-Network CNN (Fig. \ref{fig:anisotropic}). To avoid sensitivity of model orientation both networks go through a post-training data augmentation and multi-orientation pooling. Models are rotated by several random azimuth and elevation parameters. The trained CNN is decomposed into lower layers and higher layers. Each augmented input is passed into a separate CNN and passed through the lower layers independently. Each feature map is subsequently passed into an orientation pooling layer to aggregate the multi-orientation feature maps. The resulting feature map is then passed through the CNN higher layers. All training weights are initialised from the original CNN.

With exception of \cite{HackelEtAlT2017}, all methods described in this section classify a pre-segmented point cloud. \cite{HuangYouJ2016} propose a method where each point is parsed through a dense voxel grid, generating a binary occupancy grid centred at a sub sample of key points. The key points are randomly generated, although measures are applied to ensure key points are balanced over the object classes. The label of each point inside the voxel grid of size $20\times20\times20$ votes for the overall class of the grid. The voxel grid is passed into a simple 3D-CNN based on the LeNet architecture. The network consists of two convolutional layers, two max pooling layers and a fully connected layer. \cite{TchapmiEtAlL2017} demonstrated semantic segmentation could be refined using 3D-CNNs, tri-linear interpolation and 3D fully connected CRFs. Coarse voxel predictions are extracted from a 3D-FCNN made up of three residual layers sandwiched by two convolutional layers and max pooling throughout. To improve the resolution of the predictions the voxels are passed into a tri-linear interpolation function. As running a CRF directly on fine voxels is computationally inefficient, the authors argue it is better run the CRF directly on the point cloud where each point is a node. Tri-linear interpolation is therefore used to map the voxel grid coarse predictions onto the point cloud. Once transferred a CRF can be run to optimise the predictions. In doing this the FC-CRF enforces global consistency and provides fine grained semantics segmentation of the points.

\section{Multi-view CNNs}\label{multi-view}

\begin{figure}[ht!]
    \centering
	\includegraphics[width=0.75\textwidth]{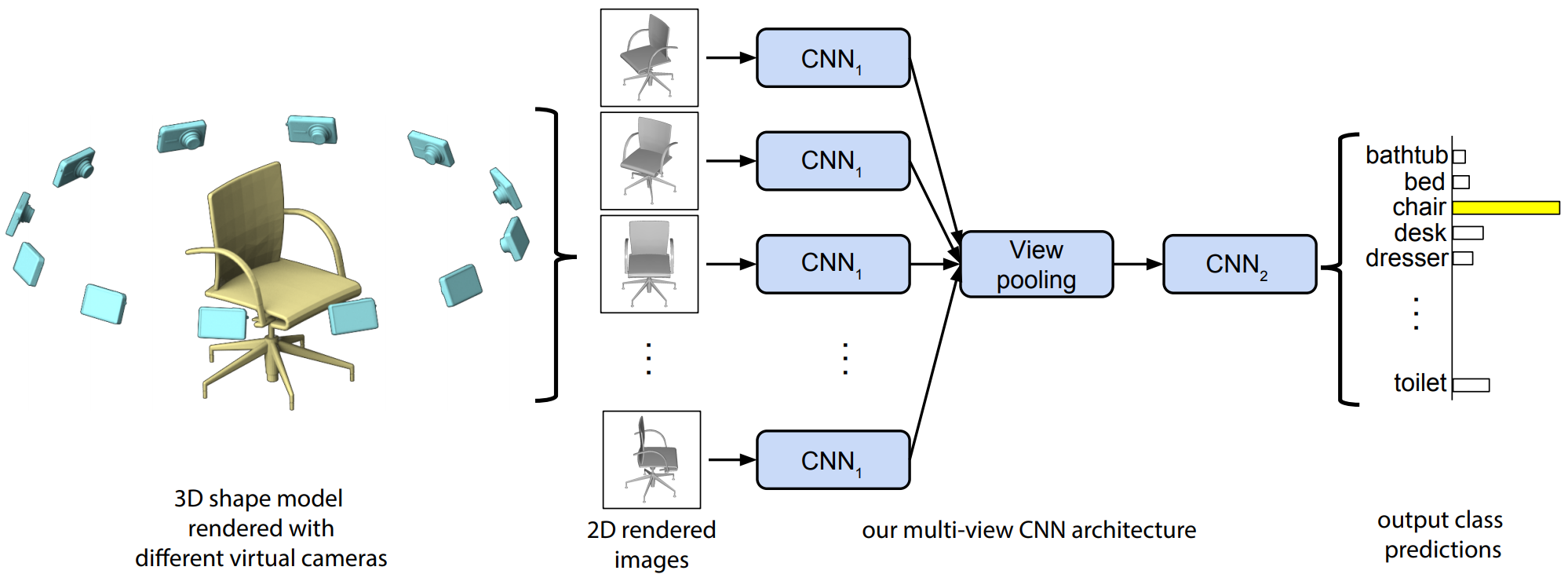}
	\caption{Multi-view CNN. Rendered 2D images are acquired with virtual cameras and fed into independent CNNs (CNN$_{1}$). Feature maps are aggregated into a single feature map describing a 3D object. This is passed into a second CNN (CNN$_{2}$) for classification. Image source: \cite{SuEtAlH2015}.}
	\label{fig:multi_view_cnn}
\end{figure}

It may seem intuitive that the best way to build 3D model descriptors is directly on 3D data. However, multi-view CNNs provide a counter argument to this. Multi-view CNNs are comprised of multiple 2D rendered views of a 3D object using standard graphics rendering to generate the view points. There are a number of reasons why multi-view CNNs perform better than volumetric CNNs, however, the most prominent are: the ability to make use of an abundance of large training datasets, decreased memory requirements allowing for higher resolution data and the ability to exploit mature 2D CNN architectures. Multi-view CNNs were proposed in the seminal paper by \cite{SuEtAlH2015}, although, previous work had been undertaken by \cite{LeCunEtAlY2004} in the use of binocular images for CNN image classification. The main novel contribution by \cite{SuEtAlH2015} is the \textit{view pooling} layer, a mechanism for combining multiple 2D renderings from a 3D object in a single end-to-end network. The network takes 12 rendered images in its simplest pipeline, and passes each image through respective independent CNNs. Each network feature map is then concatenated in the \textit{view pooling layer}, which is subsequently the input to a second aggregated CNN (Fig. \ref{fig:multi_view_cnn}). Two camera setups were proposed. The first assumes that the object is orientated upright. The 12 views can be rendered by placing virtual camera around the mesh at \ang{30} intervals. A second set up does not make use of this assumption. Instead 20 virtual cameras are placed at the 20 vertices of an icosahedron enclosing the shape, where four images are acquired at \ang{0}, \ang{90}, \ang{180} and \ang{270} rotation along the axis passing through the camera and the object centroid. A VGG-M network architecture is used to learn the features. The VGG-M network consists of five convolutional layers and three fully connected. Although the \textit{view pooling} layer can be included in multiple positions in the network, the most optimum as noted by the authors is after the fifth convolutional layer. The highest score reported by the authors using the multi-view CNN network on the ModelNet40 dataset is 90.1\%.

\cite{HeEtAlK2016} demonstrated that multi-view CNNs could be further advanced with an improved data augmentation. A multi-resolution 3D filter was proposed to capture information at multiple scales. This is achieved by performing sphere rendering at multiple volumes resolutions. Spheres are used for discretisation as they are view-invariant shapes. By doing this potential noise and irregularities are regularised making the approach more suitable for real world data. This achieved a classification accuracy of 93.8\% on the ModelNet40 (2.7\%+ from the original implementation). Again, it should be emphasised these methods output a classification score, and therefore, do not spatially correlate the accumulated features. \cite{KalogerakisEtAlE2017} proposed a method to achieve semantic part segmentation using image based FCNs \cite{LongJ2015} and CRF's. The initial layers of the network functions similar to a typical multi-view CNN with multi-scale rendering. The images are two channel images $512\times512$ consisting of a rendered shaded image and the corresponding 2.5-D depth map, which are passed into an adapted VGG-16 Net for feature extraction. The feature maps are subsequently passed through multiple convolutional transpose operators to produce a $512\times512$ confidence map of the input image which is then converted into probabilities through a CRF layer. To aggregate confidence maps from multiple views and project the results back onto a 3D surface, the authors propose the \textit{Image2Surface} projection layer. The layer takes the confidence maps along with a surface reference (polygon ID) which is presented as a 3D image $512\times512\times k$ where $k$ is the number of classes.

The above mentioned architectures rely on a full 3D mesh to enable reliable multi-view rendering. In large outdoor scenes such meshes are typically unavailable. 3D structure of outdoor scenes is commonly generated from photogrammetric principles. A common method is to first semantically segment in the 2D image space and project the pixel classes onto the 3D mesh \cite{LadickyEtAlL2010, TigheLazebnikJ2010}. Each pixel in 2D space becomes a 3D point in space which, if desired, subsequently becomes a vertex for a mesh. By a means of voting, the semantic label of each face in the mesh is computed. By definition, there is a inherent redundancy amongst the overlapping images, and therefore, semantically segmenting each image is inefficient and less effective. \cite{RiemenschneiderEtAlH2014} propose a method to exploit the 3D mesh geometry to determine optimum images that best cover a 3D area with minimum redundancy. This is achieved using a method comparable with optimum image selection in texture mapping problems. The results concluded that optimising a subset of images for segmentation not only increases compute efficiency but also has a positive impact of overall class prediction accuracy. Alternatively, \cite{qin2018deep} present a method for terrain classification for ALS point clouds using multi-view methods. First, low level features are extracted directly from the ALS point cloud. These consist of elevation, slope angle and intensity. For each viewing angle, the point cloud is projected on to an image plane. Grid interpolations are then performed over the projected point cloud generating multimodal representations for each respective viewing angle. The resulting network TLFnet demonstrates to our knowledge the first general method for applying multi-view CNNs to large-scale aerial point cloud segmentation.

\cite{DaiNiessnerA2018} present a sophisticated approach for indoor semantic scene segmentation and modelling that exploits depth map geometry and spectral data from RGB-D cameras. Although, strictly a volumetric network architecture, multi-view aggregation is achieved with a multi-view pooling layer to handle varying numbers of RGB input views of the 3D representations. The pipeline first extracts 2D feature maps from RGB images which are downsampled through multiple convolution layers to generate rich feature maps. The 2D feature maps are backprojected into 3D space where a 3D CNN learns both the backprojected 3D image features as well as direct 3D geometric features. The 3D information comprises of chunks of $31\times31\times62$ voxels for a given $x,y$ location in the scene. This equates to a spatial neighbourhood of $1.5m\times1.5m\times3m$ in height. A key component and novel contribution of the network is the \textit{differentiable backprojection} layer. 2D-3D associations are computed on-the-fly by assuming the knowledge of a 6-DoF pose alignment for each RGB input image with respect to each other and the 3D reconstruction. First, voxel to pixel projections are computed based on corresponding camera pose, camera intrinsics and the world-to-grid transformation matrix. This enables a per-point voxel-to-pixel mapping which is used to project the last layer of the 2D CNN onto the voxel grid used as the first layer for the 3D-CNN. This work demonstrated that by aggregating spectral information with voxelised 3D data a 22.2\% performance increase could be realised on the ScanNet 3D benchmark, in comparison to the existing state-of-the-art. This is a significant result as it demonstrates the combination of both data types does not lead to redundancy, but instead leads to performance increase.

\section{Unordered point set processing}\label{unordered}

\subsection{Supervised}

The prior deep learning approaches reviewed in this paper all have a common assumption; the input data is in a regularised data structure with explicit connectivity information. This is achieved by either volumetrically discretising an unordered point cloud (i.e. voxel grid) or by representing the 3D data as an ordered 2D image array (RGB-D and multi-view). Whilst these methods achieve very good results in classification, object detection and segmentation tasks, it remains desirable to find network architectures that can directly consume unordered point clouds. Such networks can provide end-to-end classification, object detection or segmentation without the memory overheads of voxel grids or the potential loss of information from 2D image representations.

\begin{figure}[ht!]
    \centering
	\includegraphics[width=\textwidth]{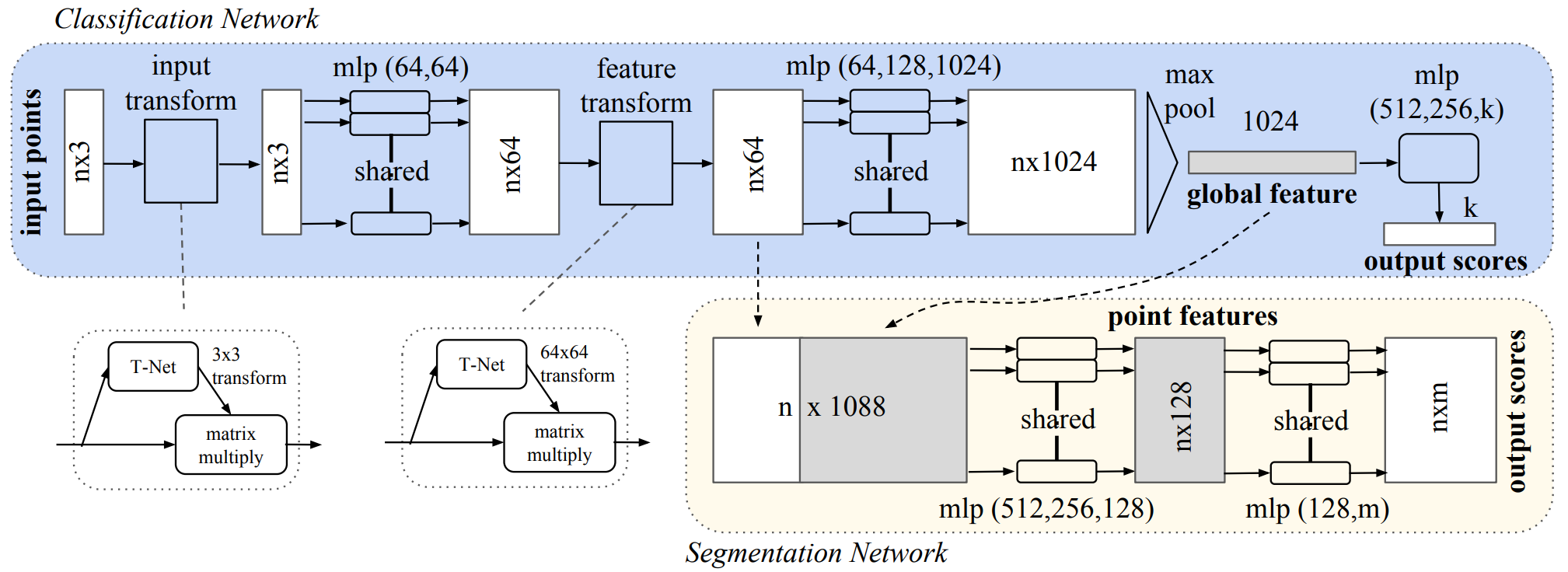}
	\caption{PointNet classification and segmentation network architectures. The network consumes raw $x,y,z$ point cloud data of size $n\times3$. T-Net is implemented to learn input permutations and canonocalise the data. Global features can be passed into a segmentation network where they are concatenated with point features to combine global semantics with local geometry. All layers consist of a \textit{batchnorm} and \textit{ReLU} operation and \textit{dropout} is used for the final MLP layer. Image source: \cite{QiEtAlC2017b}.}
	\label{fig:pointnet}
\end{figure}

\cite{QiEtAlC2017b} proposed the first network fulfilling this criteria with their seminal architecture PointNet. PointNet differs significantly from other architectures discussed in this paper in that it does not extract features with convolution operators, but instead, consists only of fully connected layers. In its simplest form each point $\mathbf{P}\in\mathbb{R}^{3}$ is represented by its $x,y,z$ coordinates in a Euclidean space \footnote{More complex features could include normals, eigenvectors etc. however, have not been formally evaluated.}. For classification purposes, the network aims to learn a spatial encoding for each point which are aggregated into a global point cloud signature. Features are generated using multi-layer perceptrons (MLPs) and aggregated using a single symmetric function, max pooling (Fig. \ref{fig:pointnet}). In essence, the network learns a set of functions that select interesting and informative key points from a subset of points, encoding this information in each layers feature vector. PointNet also offers a segmentation sub network which concatenates the aggregated global features extracted using the classification sub network and passes the features into two MLPs to generate per-point features and subsequently class probabilities for each point. Per-point features are obtained by concatenating global and local feature vectors. Using this method, the authors demonstrated the networks ability to learn rich local features by accurately predicting per-point normal values.

To aid classification it is important to present the objects orientation in a canonical form to increase invariance to input permutation. As both rigid and affine transformations are simple to apply to point clouds, a small sub-net \textit{t-net} is implemented into PointNet to attempt to transform the data into canonical form. This is applied to both the input point cloud as well as to the feature vector after the first MLP operator. A further strategy to increase input permutation invariance is to also apply small permutations to the training data (i.e. jittering).

\begin{figure}[ht!]
    \centering
	\includegraphics[width=0.75\textwidth]{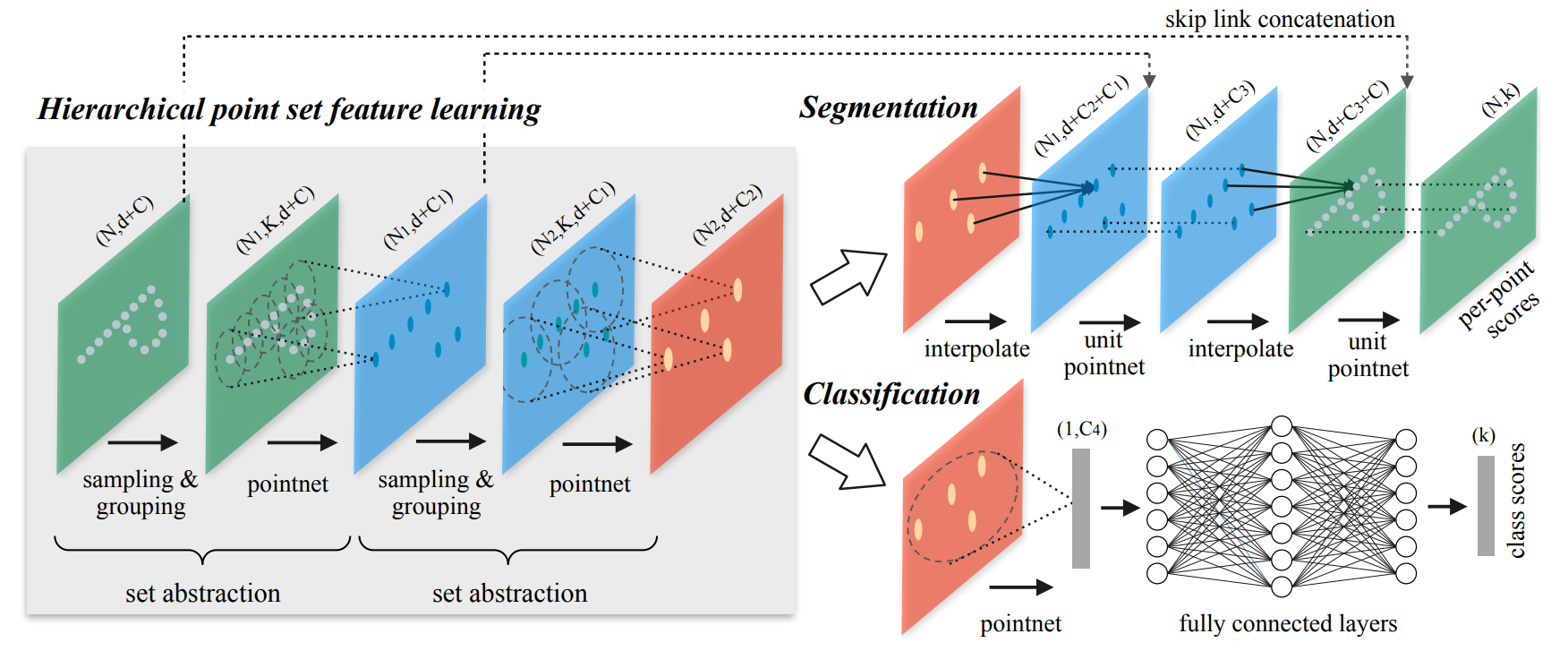}
	\caption{PointNet++ architecture. Hierarchical feature learning is introduced to learn features at various scales. Sampling and grouping layers define and sample neighbourhoods at various sizes which are fed into a mini-PointNet architecture for feature extraction. Image source: \cite{QiEtAlC2017a}.}
	\label{fig:pointnet++}
\end{figure}

A key issue with PointNet is that, by design, it does not capture local structures induced by the metric space points occupy. This causes limitations concerned with fine grained local patterns. To rectify this \cite{QiEtAlC2017a} proposed PointNet++, a hierarchical neural network. PointNet++ takes inspiration from 2D-CNNs where inputs capture features at progressively larger scales along a multi resolution hierarchy. Point sets are partitioned into overlapping local regions by a distance metric. Features are then extracted from a progressively increasing neighbourhood size. Whereas small neighbourhoods capture fine grain local features (i.e. surface texture), large neighbourhoods capture global shape geometry features. The original PointNet architecture (Fig. \ref{fig:pointnet}) is used for feature extraction. To generate overlapping partitions a neighbourhood ball is defined where a given point is used as the centroid location and radius dependant on the hierarchical scale parameter. The \textit{farthest point sampling} algorithm is used to ensure even coverage of the whole point set. The authors describe the process as three layers; the \textit{sampling} layer defines a point set from the global point set, a \textit{grouping} layer constructs the local region of points within the defined sample, finally the \textit{PointNet} layer uses a mini-PointNet to extract local region features as feature vectors (Fig. \ref{fig:pointnet++}). When selecting points by neighbourhood size, point density plays a significant role. Unlike images, where an image is a uniform grid, point cloud density can vary across a scene, meaning uniform density cannot be assumed. PointNet++ demonstrates that unlike in 2D CNNs where small kernels are preferred, when point density is sparse, larger point samples are required for robust pattern extraction. To learn an optimised strategy to combine multi-scale features, a \textit{random input dropout} layer is introduced. This layer randomly drops out input points with randomised probability for each instance at a dropout ratio $\theta$ uniformly. This allows the network to train with varying sparsity, encouraging invariance against differing point density. PointNet++ achieved a benchmark score of 90.7\% on the ModelNet40, a 2.7\% increase from PointNet.

Taking a similar approach to PointNet++, \cite{Engelmann2018} use a combination of $K$ Nearest Neighbours (KNN) and $k$-means to add spatial pooling to a MLP network. The proposed network consists of feature blocks where each feature block passes all points through a MLP then compute a global aggregation by max pooling. The global feature is then subsequently passed through an MLP. These blocks can be stacked to arbitrary depths. Two grouping methods are used to obtain spatial feature information. The local feature space for each point $p$ is derived by computing an aggregation of the KNN of points, generating a KNN-tensor, in a Euclidean space. Pooling is achieved through a $k$-means approach, where $k$ represents to number of points after the pooling. Each point is represented as the centroid of each cluster where the feature is an average of all points within the cluster. Additionally, two loss functions are proposed; $\mathcal{L}_{pair}$ and $\mathcal{L}_{cent}$. The first, $\mathcal{L}_{pair}$, enforces the assumption that semantic classes are likely to be nearby in the feature space and is a direct measurement of this semantic similarity. This means the network will learn an embedding where two points of the same class are nearby in the feature space. The second, $\mathcal{L}_{cent}$, reduces the distance of a point feature $x_i$ and the corresponding centroid $\bar{x_i}$. Here the cosine distance was found to be more effective than both $L_1$ and $L_2$ distance measures. This relatively simple extension of PointNet achieved mIoU mean class Intersection over Union (mIoU) of 58.3 and 25.39 for S3DIS and ScanNet respectively. 

PointSIFT \cite{JiangEtAlM2018} propose a 'SIFT-like' module that can be implemented in PointNet style networks. PointSIFT is inspired by the Scale Invariance Feature Transform (SIFT) \cite{LoweD2004} feature descriptor. Similar to SIFT, the module attempts to encode information of different orientations and is adaptive to scale. Orientation is encoded by convolving the features of the nearest points in eight orientations. As opposed to using $k$ nearest neighbours as in PointNet++ information is captured for all points in the local neighbourhood. By incorporating the PointSIFT module on the PointNet++ architecture the authors achieved a segmentation accuracy of 86.2\% overall accuracy and an 8.4\% increase in IoU.

Since PointNet(++) there has been an influx in research operating directly on unordered point clouds. In particular there have been significant effort to incorporate a spatial convolution operator within the network. A key example of this is SplatNet \cite{Su_2018_CVPR}. SPLATNet (SParse LAT-tice Network) takes inspiration from the permutohedral lattice \cite{AdamsEtAlA2010} where convolutions are performed on sparse data in high dimensions. This is done efficiently by using sparse bilateral convolutional layers, which use indexing structures to apply convolutions only on occupied parts of the lattice. A key difference to PointNet++ is that max pooling layers are not used to aggregate information through multiple scales. Instead, flexible specifications of the lattice structure are designed to enable hierarchical and spatially aware feature learning. In essence, points are mapped onto Bilateral Convolutional Layers (BCL) using a barycentric interpolation. Once convolutions are performed in this higher dimensional lattice, the results then undergo barycentric interpolation to be mapped back to the original points. These three processes are referred to as; Splat, Convolve and Slice respectively. Also unique is the ability to combine 2D features from corresponding images to strengthen features. This is particularly useful for applications such as photogrammetry and MLS where 3D point clouds and corresponding registered 2D images are available. Features for a corresponding 2D image is first past through a CNN feature extractor prior to a BCL layer. The BCL then maps 2D features onto a 3D lattice where after 2D and 3D features can be concatenated. The joint features are further passed through two 1x1 convolutional layers where finally a softmax layer enables point wise class probabilities. SPLATNet achieved a mean IoU of 83.7 on the ShapeNet part-segmentation benchmark. Further experiments on the RueMonge2014 \cite{RiemenschneiderEtAlH2014} fa{\c c}ade benchmark dataset yielded an average IoU score of 0.654 for 3D point features only, 0.698 with 2D-3D features, however the best results were realised with a multi-view 2D image approach (70.6).

A prominent issue with point cloud classification is the lack of adequate high quality training data. Despite the datasets reviewed in \textit{Section \ref{datasets}} these still do not compare to the size of 2D image datasets such as ImageNet. SqueezeSeg \cite{WuEtAlB2018a} and SqueezeSegV2 \cite{WuEtAlB2018} takes a similar approach to SPLATNet in that 3D points are mapped onto into a more suitable state for a convolution operator. Here, 3D points are projected on a spherical surface, where a standard 2D convolutional operator is performed. Further refinement of the point-wise classification is performed by a CRF which is implemented as a recurrent layer. A key difference in SqueezeNet(V2) is that the model trained using synthetic training data. This is acquired by passing a virtual Velodyne sensor through the large virtual environment of the popular game Grand Theft Auto V. SqueezeNet generalised very poorly to real data however, with a test accuracy of 29\%. SqueezeNet V2 accredits the poor generalisation to \textit{dropout noise}. The authors define this as missing points from the sensed point cloud caused by limited sensing range, mirror diffusion of the sensing laser, or jitter in the incident angles. SqueezeSeg V2 proposes a domain adaption training pipeline. Firstly, dropout noise is mitigated by the novel Context Aggregation Module, which aggregates contextual information from a larger receptive field. Other improvements include regularisation techniques such as batch normalisation and focal loss. Finally, a occupancy binary mask layer is added where 0 denotes no point present, and one point is present. By reducing dropout noise, adding regularisation and using the occupancy binary mask as an input channel, test accuracy on real world datasets increase to 57.4\%.

Another key challenge in raw point cloud processing is the natural non-uniform distribution of real world data. This can occur from occlusions, distance from sensor and sensor noise to name just a few. These characteristics mean applying a spatial convolution is very challenging. \cite{HermosillaEtAlP2018a} address this by proposing a novel method that first represents the convolution kernel as a multilayer perceptron, next phrases the convolution as a Monte Carlo integration problem and lastly, utilises Poisson disk sampling as a scalable means of hierarchical feature learning. The authors demonstrate by estimating the convolutional integral with Monte Carlo computation, with proper handling of the variance in underlying point sampling density, state-of-the-art performance can be achieved for model and point-wise classification. By implementing this approach, the network gains a level of sampling invariance, whereby the convolution becomes invariant to both the order of the points and the variable number of neighbours for each sampled point. Furthermore, the use of Poisson disk sampling for point hierarchy construction (as opposed to the more commonly used farthest point sampling using in PointNet++) demonstrates a higher level of scalability and allows to bound the maximal number of samples in a receptive field. The network achieves 85.9\% segmentation accuracy on the ScanNet uniform test dataset. However, perhaps more importantly, achieves 85.2\% on the non-uniform test dataset. In a similar setting PointNet++ accuracy decreases ~4\% when classifying non-uniformly sampled point clouds.

PointCNN \cite{LiEtAlY2018a} also uses convolutions directly on the point cloud. Here, a $k$ nearest neighbours is used to find spatially local points to introduce point order invariance. The network uses an MLP on the derived local point neighbourhood and learns a transformation $\mathcal{X}$ of size $k\times k$ on points $\{p_{1},...,p_{k}\}$, which is used to weight and permute the input point features. The latter has a similar affect of PointNet's \textit{T-Net} which attempts to rotate the point cloud into canonical order. Convolutions are subsequently applied on the $\mathcal{X}$-transformed features. PointCNN further expresses the importance for hierarchical feature representations for effective point cloud classification. Although conceptually simple, PointCNN achieves 85.1\% on the ScanNet uniform benchmark dataset. In a similar attempt to address the lack of spatial convolution \cite{thomas2019kpconv} present KPConv offering a deformable convolution operator. Each local neighbourhood is convolved by applying the weights of the nearest distance kernel point in the neighbourhood. Originally these are set uniformally, which in essence is a pseudo local voxelisation. However, the position of the kernel points are also learned in the network allowing the points to learn the topology of the local neighbourhoods and deform the voxel grid to suit the such a topology. Similar to \cite{HermosillaEtAlP2018a}, poission disk sampling is used to sub sample the points and a radius search is performed to gain the local neighbourhoods for pooling. This method achieved mIoU scores of 68.4, 73.1, 67.1 and 75.9 mIoU for ScanNet, Semantic3D, s3DIS and MPM3D respectively. 

\begin{figure}[ht!]
    \centering
	\includegraphics[width=0.75\textwidth]{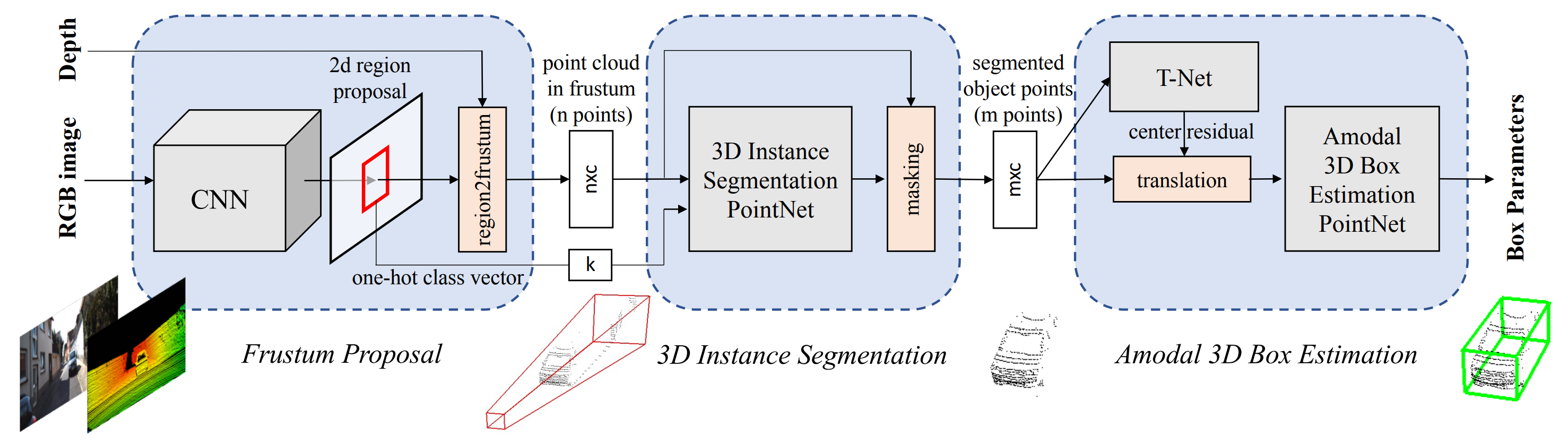}
	\caption{Frustum point net architecture. 2D CNN object detection is used to determine objects in RGB-D depth maps. Frustums are lifted from the 2D image to create a 3D point cloud search space. A PointNet instance segmentation network performs a binary segmentation for the predicted class (encoded in the one-hot-class vector). A \textit{t-net} is used to refine the segmented points centroid to align close to ground truth bounding box centroid. Finally, a 3D amodal bounding box estimation is regressed, optimised using the novel \textit{corner loss} function. All three networks are subsequently trained using a multitask loss function encompassing all three networks. Image source: \cite{QiEtAlC2017}.}
	\label{fig:frustum_pointnet}
\end{figure}

Currently unordered point set networks are not capable of instance level object detection (with exception of \cite{WuEtAlB2018} who use an unsupervised clustering algorithm as a post processing step). An obvious approach would be to simply pass a 3D sliding window over the scene \cite{EngelckeEtAlM2017} or train a 3D region proposal network \cite{SongEtAlS2015}. However, as computational complexity of 3D search typically grows cubically with respect to resolution these methods are not efficiently scalable for large point sets. This was the motivation for Frustum PointNets \cite{QiEtAlC2017}, a 3D object detection network. The network classifies object classes and estimates oriented 3D amodal bounding boxes in both indoor and outdoor scenes. The key approach to Frustum network, as the name implies, is the generation of frustum proposals in the point cloud projected and detected from RGB-D depth images (Fig. \ref{fig:frustum_pointnet}). This not only reduces the search space by a dimension (3D to 2D), but also allows for the exploitation of mature 2D object detection networks and training datasets. Due to frustums having a range of heading orientation, given the known camera projection matrix, frustums are normalised by rotation, such that the frustum is orthogonal to the image plane which aids rotation invariance. 3D frustums act as search parameters and the sampled point sets are subsequently passed into a 3D instance segmentation PointNet network. As frustums represent space with only one object of interest, a binary segmentation is then carried out for all sampled points. Using the class from the 2D detector, the information is encoded as a one-hot class vector. This means the network can search for geometry related to the proposed feature class. To further refine orientation, PointNets \textit{t-net} \cite{QiEtAlC2017b} regression network estimates the true centre of the object and realigns the object centre. A 3D bounding box regression network is performed to estimate the box centre $c_{x}, c_{y}, c_{z}$, size $h, w, l$ and orientation $\theta$. All three networks are trained simultaneously using the network architecture visualised in \ref{fig:frustum_pointnet}. Training is optimised with a multitask loss algorithm which encompasses loss from all three networks, to encourage a uniform optimisation. The frustum network demonstrated state-of-the-art performance on a range of benchmark datasets including 54.0 mAP on the SUN-RGBD and obtaining leading results on KITTI 3D object detection and localisation by considerable margins.

An alternative to view frustum projections is the ability to regress the 3D boxes directly in the 3D space.  \cite{yang2019learning} propose a methodology for this in their network 3D-BoNet. Both 3D bounding box regression and point level masks for each instance are performed simultaneously. A key advantage of 3D-BoNet is the efficiency of the algorithm. Both in network training and with the eliminaiton of pre/post processing required for Frustum PointNets. 3D-BoNet further achieves state-of-the-art on the S3DIS instance segmentation dataset with a mean precision and recall of 65.6 and 47.6 respectively. 

\subsection{Unsupervised}

Advancements of sensors and software capable of generating 3D point cloud now make it trivial and cheap (both monetary and computationally) to collect very large point cloud datasets. As discussed, manually labelling this data for supervised learning is both technically challenging and laborious. It would therefore be favourable if algorithms could learn from this wealth of data without the need for manual labelling. Such an approach of learning is generally referred to as either unsupervised or self-supervised. A typical approach for learning structure is through auto-encoders. Simply, an auto-encoder is a network that attempts recreate its input as an output. Formally, given an encoder $\textbf{h}=f(x)$ where $\textbf{h}$ is a hidden layer (or set of), and a decoder that produces the reconstruction $\textbf{r}=g(\textbf{h})$ the auto-encoder loss function $L$ attempts to minimise $L(\textbf{x},g(f(\textbf{x})))$ where $L$ is penalised for $g(f(x))$ being dissimilar from $\textbf{x}$ \cite{GoodfellowEtAlI2016}. However, unlike an image where dissimilarity can be compared by comparing corresponding pairs of pixels, 3D data must have a 3D dissimilarity measurement. Such an examples include the Chamfer loss (Eq. \ref{eq:chamfer_dist}) and earth-movers-distance (Eq. \ref{eq:earth_mover_dist}) \cite{RubnerEtAlY2000} for per-point distance comparison. In training an auto-encoder, the network learns to preserve the most important features for the intrinsic structure of the input point cloud, as the latent space at layer \textbf{h} is considerably smaller than the original input. In practice, once the encoder has learnt the optimum features for the training data, the decoder can be replaced with a series of fully connected layers followed by a softmax layer for classification. By using the pre-trained weights for the encoder, such a network can now be trained with considerably less data, and still benefit large wealth of unlabelled 3D data.

\begin{equation}
d_{CD}(S_{1},S_{2})=\sum_{x\in S_{1}}\min_{\substack{y\in S_{2}}}\|x-y\|_{2}^{2}+\sum_{y\in S_{2}}\min_{\substack{x\in S_{1}}}\|x-y\|_{2}^{2}
\label{eq:chamfer_dist}
\end{equation}

\begin{equation}
d_{EMD}(S_{1},S_{2})=\min_{\substack{\phi:S_{1}}\rightarrow S_{2}} \sum_{x\in S_{1}}{\|x-\phi(x)\|_{2}}
\label{eq:earth_mover_dist}
\end{equation}

\noindent where $S_{1}$ and $S_{2}$ are equal size point sets where $\textbf{S}\in \mathbb{R}^{3}$ and $\phi$ is a bijection.

\cite{LiEtAlJ2018} propose a permutation invariant network called SO-Net. A spatial distribution of the input points is built using self-organising maps. A hierarchal feature extraction is performed on both the individual points and self-organising map nodes, ultimately representing the input point cloud by a single feature vector. An adjustable receptive field is used to allow for more effective local feature aggregation by adjusting according to local geometry. Similar work by \cite{YangEtAlY2017} present FoldingNet. Here a graph based enhancement is enforced to promote local structures on top of PointNet. 3D objects are represented as point clouds by a folding based decoder that deforms a canonical 2D grid onto the underlying 3D object surface of the inputted point cloud. This method is not a fully end-to-end single network, and final 3D classification is carried out by a SVM on the latent space features. 

\begin{figure}[ht!]
    \centering
	\includegraphics[width=0.75\textwidth]{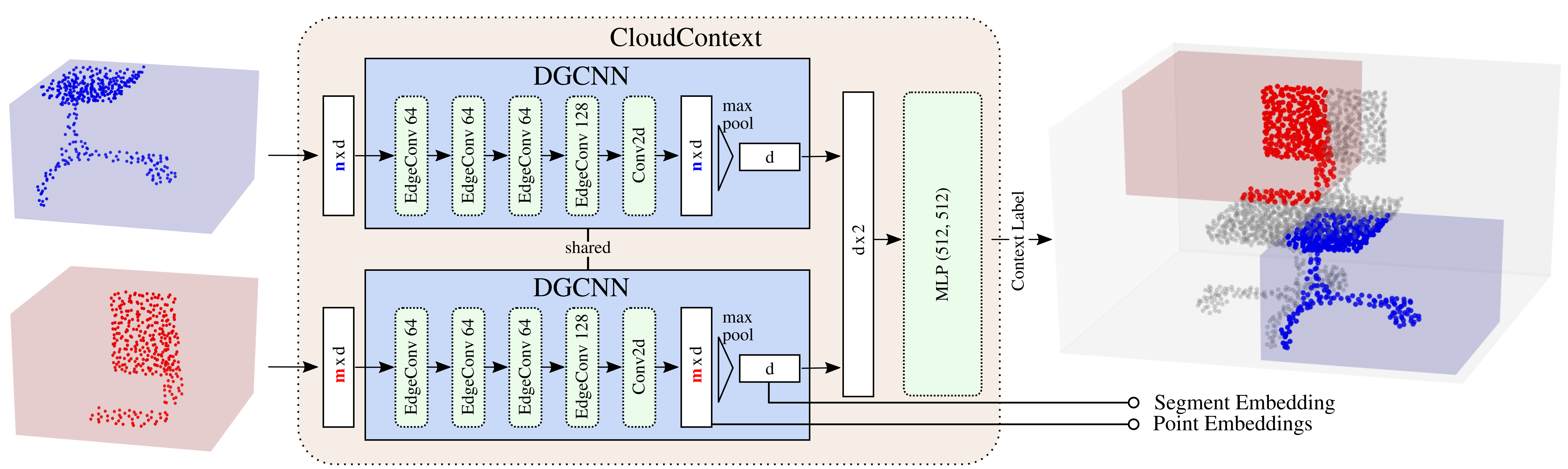}
	\caption{DGCNN network architecture for unsupervised context prediction for point clouds. The intermediate latent space feature descriptors are useful for classical machine learning model training such as Support Vector Machines and Random Forests. Image source: \cite{SauderSieversJ2019}.}
	\label{fig:dgcnn}
\end{figure}

\cite{AchlioptasEtAlP2017a} directly re-purpose the PointNet to encoder with a fully connected decoder network. An additional three fully connected layers are added onto the decoder of size 1024, 1024, $n\times3$, where $n$ is the original number of input points. Gaussian Mixed Models were also trained with within the fixed latent space of the auto-encoder which yielded the best reconstruction results on ModelNet objects. \cite{SauderSieversJ2019} extend this approach for context prediction of two segments of a continuous point cloud. In learning the contextual relationship between two point clouds the need for computationally expensive per-point measurement loss function is circumvented. This builds on prior work by \cite{DoerschEtAlC2015} who demonstrate learning how two patches for the same image correspond can be a powerful initialiser for CNNs. Instead, two parts from the same point clouds are both passed into separate feature encoders where the weights are subsequently shared Fig. \ref{fig:dgcnn}. Point-to-point loss is avoided by posing this as a classification task in 3D space.

\cite{ZamorskiEtAlM2018} extend the variational auto-encoder technique to an adversarial auto-encoder network. The proposed network 3dAAE is a fully end-to-end solution which simultaneously learns a latent space representation and generates a 3D shape from the latent code. Unlike the variational auto-encoder an extra branch containing a discriminator is added to the network. The discriminator is passed latent codes from both the encoder and from an arbitrary prior distribution. This adds an incentive for the encoder to constrain the latent space distribution to the chosen prior. In this work a normal distribution was used as the prior. The authors suggest that a Chamfer loss is more suitable then EMD as it is fully differentiable with less computational demands. The encoder is very similar to the original PointNet architecture. The authors evaluate their network on a single class (chair) of the ModelNet dataset, however, demonstrate that the introduction of a prior distribution constraint is beneficial for both dimensionally reduced representation of point cloud data and for generative seed codes.

\section{Ordered point cloud processing}

Whilst in Section \ref{unordered} we looked at algorithms that directly process unordered point clouds, this arguably underestimates the whole research area for end-to-end learning directly on point clouds. A variety of literature is focused on ordering point clouds for processing for classification and segmentation. One of the most prominent examples of this is OctNet proposed by \cite{RieglerEtAlG2017}. Here point cloud data is represented and indexed as shallow octrees. OctNets are modified CNNs that are capable of operating on non-uniform grids. This is achieved by first hierarchically partitioning the point cloud space using a set of unbalanced octrees where each leaf node stores a pooled feature representation. Data is split according to local density of the data, so that areas of high density undergo more splits. Convolution operators are then directly defined on the structure of the trees as opposed to the input data itself. OctNets demonstrate the ability for convolution and pooling operations to be implemented on this type of data structure. Similar to OctNets, \cite{KlokovLempitskyR2017} index point clouds using a kd-tree structure. A deep network is then constructed that operates directly on the kd-tree structure. By representing the data as kd-trees over octrees a 4.7\% performance increase was realised on the Model40 benchmark dataset. As with a CNN the network is a feed forward but where the learnable parameters are associated with the weights of the nodes in the kd-tree. Kd-Network also demonstrated an efficiency in both training and inference, suggesting scalability. These methods are often used for non-euclidean data such as social graph data such as interactions between users on social network websites. For a full review looking into graph based deep learning for processing non-euclidean unstructured data the reader is referred to \cite{bronstein2017geometric}.

\cite{LandrieuSimonovskyL2017} demonstrate an effective graph based method, where 3D point clouds are organised in a novel structure known as a \textit{superpoint graph} (SPG). SPGs are derived by partitioning the point cloud into geometrically homogeneous elements. The graph is a collection of interconnected simple shapes referred to as super points (inspired by 2D super pixels). Each node in the graph represents a simple shape while edges describe their adjacency relationship characterised by rich edge features. The authors argue by representing data as SPGs aids classification and segmentation as whole objects are classified and segmented as opposed to individual points/voxels. To train the model for classification the PointNet architecture is used for superpoint embedding and graph convolutions are used for contextual segmentation. The graph convolution method used is an extension of Edge Conditioned Convolutions \cite{SimonovskyKomodakisM2017}. The proposed network achieved a state-of-the-art score on the Semantic3D TLS benchmark dataset with an overall accuracy of 92.9\%, an increase of 10.4\% to PointNet++. Further Graph CNNs have been proposed using either spatial filtering \cite{ScarselliEtAlF2009} or spectral filtering \cite{BrunaEtAlJ2013}, however, these are generally concerned with meshes or point clouds with a known intrinsic structure. As such we do not discuss these methods here although, a review has been undertaken by \cite{AhmedEtAlE2018}.

\section{Discussion}

In the previous sections we have reviewed the current prominent techniques for 3D deep learning. Although for specific benchmark datasets quantitative comparison is possible, such an evaluation approach is not applicable for a general understanding. Instead, each approaches and even individual architectures can be more applicable for some tasks but not others. Variables such as data, run time speed, computational resources and required output all dictate the most suitable approach. In this section we will look at advantages and disadvantages of each method, highlighting where one is more applicable then others. Furthermore, we look at alternative ways of representing datasets resulting in alternative approaches.

Arguably, the simplest method to extract feature maps from 3D data is to represent the data as a single 2D RGB-D image. This not only allows exploitation of large 2D training datasets, but also the maturity of 2D-CNN architectures. Moreover, processing in 2D is substantially more computationally efficient. This therefore makes the processing fast and scalable with limited preprocessing requirements. The main criticism of RGB-D is that geometric information is inevitably lost during dimensionality reduction. Furthermore, this can also make integration of information across viewpoints impractical, or in many cases impossible. However, the ability for quick processing of high resolution images makes this method one of the most suitable for geographically large datasets such as airborne photogrammetric/lidar acquired point clouds or real time robotic applications. Moreover, by converting depth data into HHA space, further information can be gained during the learning process. Perhaps the most attractive aspect of RGB-D is the ease of data acquisition owed to the advent of low cost RGB-D sensors. Although these sensors only work reliably in indoor environments, the ability to record RGB-D in real time makes this data type ideal for indoor robotic navigation and \textit{pick-and-place} applications. This is obvious from the literature where most papers focus on these application.

Volumetric approaches were developed with the key principle of scaling 2D-CNNs into 3D with the 3D convolution operator. Whilst such methods offer potentially good results, these are generally limited to small datasets or in most of the reviewed literature, single object classification. The reason for this is two fold; firstly, the computational overhead of high resolution voxel grids often makes volumetric approaches unpractical for the application. This is due to the large number of parameters associated with 3D CNNs. As such, voxelisation of large areas, for both density and occupancy grids are unfeasible. To counter this, it is common to reduce resolution of objects to small voxel grids (i.e. $30\times30\times30$), however, in doing this it is likely the information gain of working in 3D space is outweighed by information loss from down sampling. This was quantified by \cite{SuEtAlH2015}, and motivated the multi-view CNN architecture. In light of this, \cite{ZhiEtAlS2018} proposed a solution for reducing the computational overhead with their real time network LightNet. LightNet exploits multi tasking to learn multiple features simultaneously, resulting in a overall speed increase. Moreover, by adding batch normalisation between the convolution operator and activation operator the network demonstrated an ability to converge earlier with fewer parameters. This resulted in an increase of 24.25\% on the ModelNet40 over VoxNet with 67\% fewer parameters. The second issue is associated with object localisation in a 3D search space. The majority of literature either use sliding windows (i.e. \cite{SongXiaoS2016}) or require pre-segmented voxel grids (i.e. \cite{MaturanaSchererD2015}) as an input. Whilst segmentation is possible \cite{HackelEtAlT2017} this is considered very expensive and does not obtain state-of-the-art results over other methods discussed. Volumetric grids are therefore suited for classification of scenes where classical unsupervised segmentation techniques such as Euclidean clustering are robust and reliable for data preprocessing.

Multi-view representation is generally shown to out perform volumetric approaches with less computational demand. However, as with volumetric approaches it is not clear how to effectively employ this method on 3D point models that suffer from missing data or when objects are partially occluded/overlapping. With respect to 3D CAD models, virtual view points can be easily programmed to cover the entire object. However, this assumes the objects centroid and extent are known. It is not clear how a view aggregation approach can be best adapted to more unpredictable real world environments where such assumptions cannot be made. Further research into the number of virtual view-points and their optimum positions would be valuable to be able to confidently apply these methods in more complex indoor/outdoor scenes. However, as each view can often be processed in parallel using individual CNNs for each view, multi-view can offer quick performance increases. For this reason, even though the depth information is implicit, it is likely many researchers and industrial applications will use this approach. Multi-view CNNs have also demonstrated ability to be integrated into the point cloud generation when derived from multi-view stereo \cite{RiemenschneiderEtAlH2014}. Hence deriving semantically segmented point clouds and meshes directly.

The ability to directly consume unordered point clouds has been long sought after for 3D object classification and segmentation. PointNet(++) was the seminal network proposal that offered a state-of-the-art performance with such a network. This has resulted in a new surge of research looking to expand/improve on these methods. Taking into account results on benchmark datasets, it is evident for the task of per-point classification, raw point cloud deep learning approaches are leading in performance with respect to both classification accuracy and computational cost. A key limitation of this approach is fixed size of the input for deep neural network architectures. For a given area $n$ points can be input into the network for classification, however, as point cloud density is non-uniform there may be significantly more points within the area. A typical approach to account for this is to consider the neural network as a sparse classification and then in post-processing apply a form a interpolation to classify the remainder points. A network that can adapt to non-uniform point sampling density would be strongly beneficial in this domain. Furthermore, as with volumetric and multi-view techniques a large bottleneck is the expensive and exhaustive nature of determining discrete object localisation. This suggests for 3D object detection the most attractive solution in large 3D environments where classical pre-segmentation is either unreliable, not robust or inefficient is to first carry out object localisation in a 2D space. Recent work by \cite{QiEtAlC2017} demonstrated the potential for this by utilising object localisation in RGB-D depth maps to project frustums into the 3D scene. Such a method does however require accurate pose estimation of the sensor for reliable frustum projections. The network by design assumes a single instance per each frustum, and therefore does not perform well when multiple instances of objects are cluttered together. Furthermore, whilst this method exploits both 2D RGB-D and 3D, 3D proposal regions are proposed by a 2D feature extractor, therefore, full geometric information is not fully utilised. \cite{yang2019learning} offer a fully end-to-end 3D bounding box regression which offers a promising direction to account for these limitations.

All techniques discussed in this review are currently active areas of research. Whilst we have addressed where each technique is generally sought over others, there is by no means a clear cut winner. With respect to directly consuming point clouds PointNet offers the most straight forward end-to-end network approach however, currently does not achieve leading performance on benchmark datasets over other methods. Advances on the PointNet(++) architecture have also demonstrated very exciting performance improvements across a range of benchmark datasets. Despite this, it is still not clear cut whether it is more appropriate to process point clouds as raw $x,y,z$ data, or present the data in an ordered form such as a graph structure. Promising results comprising of SPG, octrees and kd-trees representations offer large potential for the future of point cloud processing. It should be noted that the review has generally been concerned with classification and segmentation, which was not by design. With exception of \cite{QiEtAlC2017, yang2019learning} and 2D (RGB-D) processing, there is a lack of effective object localisation algorithms. Whilst this hints at the potential for future research in the area, this may also suggest that object localisation is more suited in a 2D space.

\section{Conclusion}

In this paper we have reviewed the current state of deep learning techniques for 3D classification, object detection, and segmentation. Although 3D deep learning is a relatively new field, the papers reviewed demonstrate a fast growing and highly effective community of researchers. While 3D deep learning is not as mature as 2D deep learning, there is strong evidence that this gap is closing. Recent papers present state-of-the-art results using RGB-D, volumetric, multi-view, unordered point cloud and graph based techniques. This highlights that there is still no \textit{one-size-fits-all} approach, and that active research in the area will be highly beneficial to future applications. We have addressed the main advantages and disadvantages of each technique, highlighting when each is most suitable. Furthermore, we looked at the recent advancements in direct unordered point cloud processing which offers a simple but potentially very effective solution to point cloud classification and segmentation. We also reviewed the modern benchmark datasets used for network evaluation. Finally, we assessed the issues concerned with object localisation in a 3D search space which is currently a very open area for research. Current research suggests object localisation is most efficient and effective in 2D. Location boundaries are projected into 3D for classification/segmentation, however, this is a potential bottleneck and it does not make use of the 3D geometric information. We therefore address this as an important domain where future research will be highly beneficial for future applications.

\section*{Acknowledgements}

The work carried out in this paper was partially funded by Bentley Systems. The authors gratefully acknowledge this support.

\section*{References}
\bibliographystyle{IEEEtran}
\bibliography{references}

\end{document}